\documentclass[a4paper,fleqn]{cas-sc}

\usepackage[numbers]{natbib}

\usepackage{algorithm} 
\usepackage{algpseudocode} 
\usepackage{amsmath} 
\usepackage{pdflscape}
\usepackage{graphicx}
\usepackage{booktabs}
\usepackage{rotating}
\usepackage{pifont}
\usepackage{placeins}  

\let\oldsection\section
\renewcommand{\section}[1]{\FloatBarrier\oldsection{#1}}

\newcommand{\triup}{\textcolor{green}{$\blacktriangle$}}
\newcommand{\tridown}{\textcolor{red}{$\blacktriangledown$}}
\newcommand{\diamondfill}{\textcolor{blue}{\ding{118}}} 

\PassOptionsToPackage{table}{xcolor}

\algrenewcommand{\algorithmiccomment}[1]{\hfill{\it /* #1 */}} 

\def\tsc#1{\csdef{#1}{\textsc{\lowercase{#1}}\xspace}}
\tsc{WGM}
\tsc{QE}
\tsc{EP}
\tsc{PMS}
\tsc{BEC}
\tsc{DE}

\begin{document}
\let\WriteBookmarks\relax
\def\floatpagepagefraction{1}
\def\textpagefraction{.001}



\title [mode = title]{Structural Induced Exploration for Balanced and Scalable Multi-Robot Path Planning}

\author[1]{Zikun Guo }
\ead{gzk798412226@gmail.com}
\credit{Conceptualization, Methodology, Software, Writing - Original draft preparation, Writing - Review \& Editing}

\affiliation[1]{organization={Department of Artificial Intelligence, School of Electronics Engineering, Kyungpook National University},
 addressline={80 Daehak-ro, Buk-gu}, 
 city={Daegu},
 postcode={41544}, 
 country={Republic of Korea}}

\author[1]{Adeyinka P. Adedigba }
\ead{yinkpeace@gmail.com}
\credit{Methodology, validation, Writing - Review \& Editing}

\author[1]{Rammohan Mallipeddi }
\ead{mallipeddi.ram@gmail.com}
\credit{Supervision, Resources, Validation, Writing - Review \& Editing}
 
\author[2]{Heoncheol Lee }
\cormark[1]
\ead{hclee@kumoh.ac.kr}
\credit{Resources, Validation, Writing - Review \& Editing}
\affiliation[2]{organization={ Department of IT Convergence Engineering, Kumoh National Institute of Technology},
 city = {Gumi},
 postacode={39177}, 
 country={South Korea}}

\cortext[cor1]{Corresponding author}
\cortext[cor2]{Principal corresponding author}

\begin{abstract}
Multi-robot path planning is a fundamental yet challenging problem due to its combinatorial complexity and the need to balance global efficiency with fair task allocation among robots. Traditional swarm intelligence methods, although effective on small instances, often converge prematurely and struggle to scale to complex environments. In this work, we present a structure-induced exploration framework that integrates structural priors into the search process of the ant colony optimization (ACO). The approach leverages the spatial distribution of the task to induce a structural prior at initialization, thereby constraining the search space. The pheromone update rule is then designed to emphasize structurally meaningful connections and incorporates a load-aware objective to reconcile the total travel distance with individual robot workload. An explicit overlap suppression strategy further ensures that tasks remain distinct and balanced across the team. The proposed framework was validated on diverse benchmark scenarios covering a wide range of instance sizes and robot team configurations. The results demonstrate consistent improvements in route compactness, stability, and workload distribution compared to representative metaheuristic baselines. Beyond performance gains, the method also provides a scalable and interpretable framework that can be readily applied to logistics, surveillance, and search-and-rescue applications where reliable large-scale coordination is essential.

\end{abstract}



\maketitle

\section{Introduction}
Multi-robot path planning is a fundamental challenge in the field of computational intelligence and autonomous systems research. Efficiently completing spatially distributed tasks is crucial in various fields, including logistics, disaster response, military operations, and urban transportation systems. As robotic systems become increasingly popular in industrial and service applications, the need for algorithms that can efficiently solve large-scale path planning problems has become critical to operational success.

The complexity of multi-robot path planning stems from its combinatorial nature, which combines elements of the Traveling Salesman Problem (TSP) and task allocation challenges into a computationally intractable NP-hard problem, particularly when utilizing multiple core processors, as seen in \citep{han2025parallel}. While single robot path planning has been extensively studied with well-established solutions, the multi-robot scenario introduces additional dimensions of complexity through inter-robot constraints, resource allocation considerations, and the exponential growth of the solution space as the problem size increases. Traditional approaches often fail when scaled to realistic problem instances involving hundreds or thousands of nodes.

Existing methodologies for addressing multi-robot path planning can be broadly categorized into three approaches: (1) exact methods, which guarantee optimality but become computationally prohibitive for large problem instances; (2) heuristic approaches, which provide reasonable solutions with limited computational resources; and (3) metaheuristic algorithms, which explore the solution space through biologically inspired search mechanisms. Among metaheuristics, Ant Colony Optimization (ACO) has demonstrated particular promise due to its collective intelligence paradigm and ability to construct solutions through emergent behavior. While Ant Colony Optimization (ACO) has shown effectiveness in various combinatorial problems, its standard formulations exhibit significant limitations when scaled to high-dimensional or large-node scenarios. Specifically, traditional ACO often converges prematurely and lacks mechanisms to escape local optima, resulting in suboptimal routing solutions and poor scalability.

Graph theory offers powerful mathematical frameworks for understanding and solving complex routing problems \cite{burleigh2011contact}. As a foundational structure in graph theory, the Minimum Spanning Tree (MST) connects all nodes in a graph with minimal total edge cost, offering a computationally efficient and theoretically grounded approximation of the lower bound in many routing and tour optimization problems. Christofides' algorithm, which extends MST with a perfect matching, creates approximation guarantees for the TSP, offering a promising foundation for more complex multi-robot scenarios. Despite these theoretical advantages, the integration of graph-theoretic constructs with metaheuristic approaches remains underexplored in the multi-robot path planning literature.

Recent advances in hybrid algorithms \citep{jiang2010hybrid, umarani2025hybrid} have demonstrated superior performance by combining the complementary strengths of different methodologies. These approaches leverage the global search capabilities of metaheuristics while incorporating problem-specific knowledge through mathematical structures. However, existing hybrid algorithms often fail to maintain solution quality across varying problem scales or require extensive parameter tuning when deployed in different operational contexts.
To address these limitations, we propose Structural INduced Exploration (SINE), which embeds a learned/constructed graph skeleton as a persistent prior inside the pheromone based ACO. A backbone constructed from the input geometry (instantiated in this work via a minimum cost spanning skeleton) shapes both initialization and ongoing search through a persistent bias, while the ant based mechanism provides fine grained exploration and exploitation. Our approach initializes high quality tours via a backbone assisted Christofides scheme and then refines them with adaptive reinforcement that emphasizes structurally meaningful connections. This integration yields a method that (1) scales to thousands of nodes; (2) maintains balanced workload while minimizing total and maximum per robot distance; (3) remains robust across heterogeneous spatial layouts and team sizes; and (4) reduces computation through a constrained, structure aware search. Key innovations include a persistent backbone influenced pheromone deposition, an adaptive heuristic that evolves over iterations, and a load aware objective that improves fairness.

In this paper, we provide a comprehensive evaluation of SINE in multi robot path planning, comparing it against with state of the art algorithms across problem instances and involving teams of multiple robots. We analyze performance in terms of solution quality, computational efficiency, scalability, and robot workload balance. Our results demonstrate that SINE consistently outperforms existing approaches, particularly in large scale multi robot coordination tasks, where it achieves improvement in path length optimization while significantly reducing the imbalance in individual robot routes. Moreover, we provide theoretical analyses of convergence properties and approximation guarantees, establishing a rigorous foundation for its empirical success in collaborative robot systems.

\section{Related work} \label{sec:rel_wks}
Ant Colony Optimization (ACO) has been extensively studied and adapted to address the increasing complexity of path planning problems in robotics and autonomous driving \citep{zhang2025improved, bell2004ant, selvi2010comparative}. ACO-based planners generate collision-free routes for multiple robots under complex constraints \cite{ali2020path}. Combining ACO with A* and Markov decision processes to achieve smooth single-robot paths. The APF–ACO approach \cite{tan2009hybrid} enhanced state transition rules (informed by artificial potential fields) and optimized pheromone updates, leading to faster convergence and smoother paths in leader–follower formations. In addition, \cite{liang2022collaborative} addressed rescue scenarios by planning on multiple prior maps: an improved ACO–GA hybrid solved a multi-map path problem for mine rescue robots, yielding reliable routes despite map uncertainty. In industrial welding, \cite{tang2024dual} formulated a dual robot cooperative welding path planning model and proposed ACO variant with dynamic state transfer and update strategies. This solver achieved significantly shorter welding paths and faster convergence than standard ACO \cite{wang2022dual}. Some method adaptations aim to enhance path safety such as \citep{ning2018tri, gheraibia2018ant, zhu2019safety}, and optimize path quality, such as \citep{shuang2011study, pham2018aco, mcwilliams2013changes, zhu2019organizational, muhlestein2014aco, burgon2019engaging}, and accelerate convergence, leading to a variety of innovative ACO variants. In task allocation and scheduling, multi-objective and adaptive ACO \citep{lopez2010automatic, xiang2017multi, fidanova2014multi, awadallah2025multi, afshar2012multi, kaur2020maco} variants balance load \citep{jabbar2018controlling, rosenthal2011aco, arora2019aco, jiang2010aco} and deadlines across heterogeneous robots. Cooperative control often leverages ACO with graph theoretic or swarm-based frameworks \citep{wang2025distributed, chen2025multi, mehra2025combining} to guide formations or consensus behaviors. Across domains, researchers introduce novel pheromone updates, multi-colony schemes, deep learning enhancements, and environment-specific heuristics. These innovations aim to accelerate convergence, avoid local optima, and incorporate real-world constraints (uncertainty, energy, precedence, robustness) into ACO-based multi-robot algorithms.

One approach is the A* Repulsive Field ACO (AR-ACO) \citep{liu2025high}, which integrates the classical A* algorithm with the concept of repulsive potential fields from artificial potential field methods. This hybrid algorithm introduces repulsive field rules to guide ants away from obstacles, ensuring safe distances and collision-free paths. AR-ACO has demonstrated significant improvements in path safety and reliability, making it particularly effective for mobile robot navigation in environments with dense obstacles. 

Another advancement is the Intelligent Enhanced Ant Colony Optimization (IEACO) \citep{li2025intelligently}, a dynamic global pheromone update mechanism that prevents premature convergence, and a multi-objective optimization perspective that simultaneously optimizes path length, safety, and other criteria. Double Level Ant Colony Optimization (DL-ACO) \citep{yang2018new} represents another significant innovation, utilizing a two-stage approach to solve complex path planning problems. In the first stage, Parallel Elite Ant Colony Optimization (PEACO) utilizes elite ants to generate an initial global path, thereby accelerating convergence. In the second stage, this path is optimized through local optimization.

Despite these advancements, existing ACO variants often face challenges in balancing solution quality, computational efficiency, and scalability, particularly in large-scale scenarios. These limitations underscore the need for further investigation of hybrid approaches that combine graph theoretic constructs with metaheuristic frameworks. Our proposed SINE algorithm addresses these gaps by leveraging a persistent structural prior (instantiated as a minimum cost spanning skeleton) to guide the ant-based search, achieving significant improvements in path quality, scalability, and computational efficiency.

\section{Methodology}
This section presents a multi-robot path planning method that integrates a persistent structural prior into the optimization process. The prior is represented as a graph backbone, constructed from the input geometry in the form of a minimum cost spanning skeleton. This structural backbone provides a template that captures global connectivity and serves as the foundation for the optimization algorithm. Within this framework, task allocation and path optimization are jointly addressed through probabilistic modeling guided by pheromones, heuristics, and the backbone bias. Concretely, we construct the structural prior by computing a spanning skeleton from the complete graph representing the problem space, which serves as a heuristic backbone for the solution with three main components:
\begin{enumerate}
 \item \textbf{Backbone Based Initialization:} A minimum cost spanning skeleton is constructed to capture the global connectivity of the problem (line 3 in Algorithm~\ref{alg:spine-init-construct}). 
       This structural prior provides the backbone for the subsequent optimization process.
 \item \textbf{Pheromone Biasing:} Higher initial pheromone values are assigned to backbone edges to guide early exploration toward promising regions (lines 4-5 in Algorithm~\ref{alg:spine-init-construct} and line 6 in Algorithm~\ref{alg:spine-update}). 
       This bias enhances exploration efficiency by guiding the search through structurally sound and low-cost connections, while still allowing stochastic divergence when better solutions are encountered.
 \item \textbf{Heuristic Guided Search:} The search process is initialized with a structural prior bias (line 5 in Algorithm~\ref{alg:spine-init-construct}), enabling faster convergence by reinforcing backbone-aligned paths while still allowing exploration through pheromone updates (lines 3-11 in Algorithm~\ref{alg:spine-update}). 
       This balances exploration and exploitation: it stabilizes convergence by reinforcing desirable paths early, while retaining the flexibility to adapt beyond the initial backbone when data suggests improvements.
\end{enumerate}

\begin{figure}[!ht]
 \centering
 \includegraphics[width=1\linewidth]{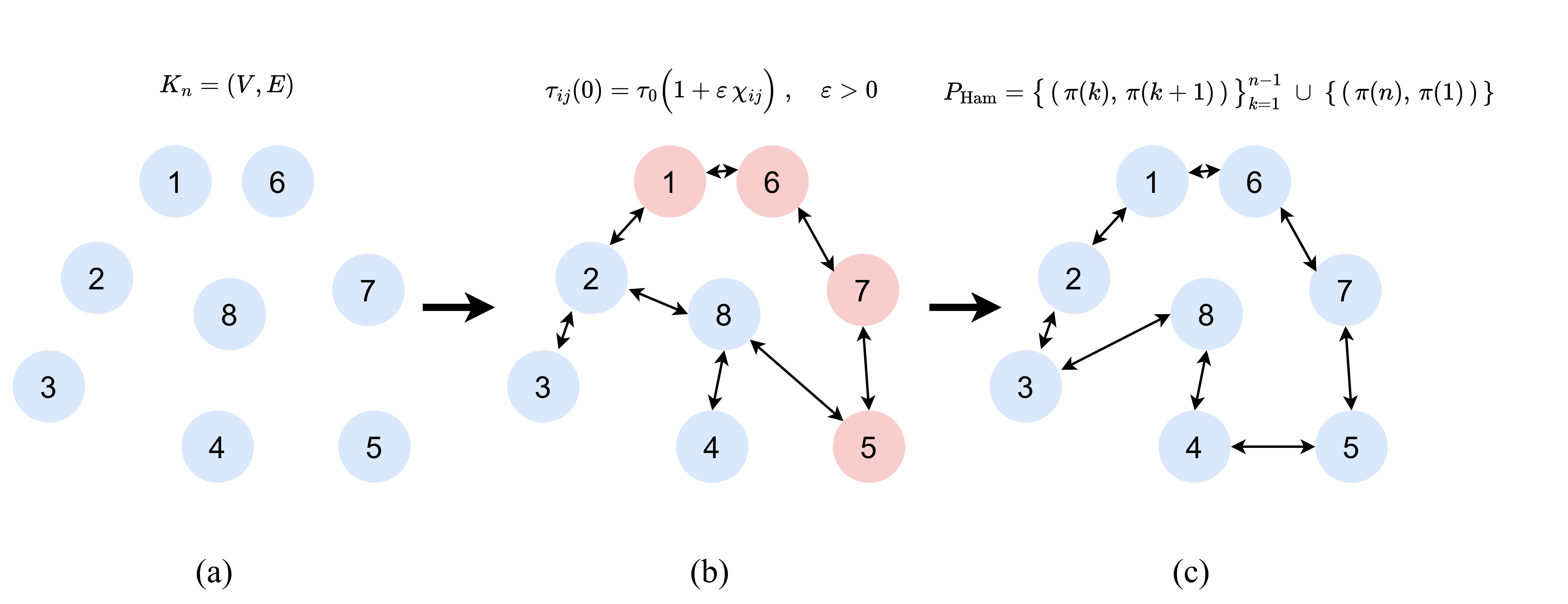}
 \caption{Three stage tour construction on the complete graph of target points: (a) the original spatial distribution of vertices prior to any heuristic; (b) the MST backbone obtained via Kruskal’s algorithm and its Eulerian skeleton produced by a depth first traversal, with provisional edges and nodes highlighted; and (c) the resulting Hamiltonian circuit after ACO refinement, in which redundant backtracks have been pruned to yield a single loop visiting each node exactly once.}
 \label{fig:original-layout}
\end{figure}

\subsection{Problem Formulation}
Figure \ref{fig:original-layout} illustrates the three-stage construction of a high-quality tour on the complete graph of eight target points. In Figure \ref{fig:original-layout} (a), only the spatial positions of vertices are shown, emphasizing the unstructured, fully connected search space prior to any heuristic intervention. Figure \ref{fig:original-layout} (b) overlays the minimum spanning tree (MST) backbone obtained via Kruskal’s algorithm, followed by a depth-first traversal that yields an Eulerian skeleton tour; red shading marks the vertices and edges that form this provisional tour, and arrows indicate the Depth First Search visitation order. Figure \ref{fig:original-layout}(c) presents the Hamiltonian circuit produced by ACO refinement: here, the redundant backtracks of the Eulerian walk have been pruned and replaced by stochastic, pheromone-guided transitions, resulting in a single loop that visits each node exactly once. Together, these figures demonstrate how the MST-based prior in Figure \ref{fig:original-layout} (b) serves as structural guidance for the ACO search, enabling rapid convergence to a near-optimal tour in Figure \ref{fig:original-layout} (c). 

Let $G=(V,E)$ be an undirected complete graph where $V=\{v_1,\dots,v_n\}$ and $E$ contains all unordered pairs $\{v_i,v_j\}$ with $i\ne j$. Each edge carries a nonnegative weight $d_{ij}=d(v_i,v_j)$ with $d_{ij}=d_{ji}$ and $d_{ii}=0$, forming the matrix $\mathbf{D}=[d_{ij}]$. Unless stated otherwise, we assume the metric (triangle inequality) condition holds. We consider a disjoint partition of tasks among $m$ robots: $V=\bigsqcup_{k=1}^m V_k$. Robot $k$ follows a tour $P_k=(v_{k1},\dots,v_{k l_k},v_{k1})$ visiting exactly the vertices of $V_k$, and its length is defined as:
\begin{equation}
L_k = \sum_{t=1}^{l_k-1} d\bigl(v_{k_t}, v_{k_{t+1}}\bigr)
 + d\bigl(v_{k_{l_k}}, v_{k_1}\bigr).
\end{equation}
In the dual objective optimization framework of multi-robot path planning, we must not only minimize the global path cost, but also consider the load balance of each path. Formally speaking, we first consider the following dual objective problem:
\begin{equation}
 \min\Bigl\{\sum_{k=1}^m L_k,\;\max_{1\le k\le m} L_k\Bigr\}.
\end{equation}
Solving this problem directly often results in multiple solutions lying on the Pareto frontier, where no single solution can be deemed globally optimal. Therefore, it becomes necessary to aggregate the multiple objectives into a unified scalar objective. In this paper, we introduce a harmonic objective function that linearly combines the total path cost and the maximum individual load through a proportional coefficient $\lambda$:
\begin{equation}
 J \;=\; \lambda \sum_{k=1}^m L_k \;+\;(1-\lambda)\,\max_{1\le k\le m}L_k,
 \quad \lambda\in[0,1].
\end{equation}

The objective exhibits the following properties. When $\lambda=1$, the objective $J$ reduces to the total path cost, expressed as $\sum_{k}L_k$, which corresponds to minimizing the overall travel effort of all robots. In contrast, when $\lambda=0$, $J$ degenerates into minimizing the worst path length $\max_k L_k$. For $\lambda \in (0,1)$, $J$ provides a continuously tunable trade-off between global efficiency and load balancing.

Furthermore, under a given set of fixed path allocations $\{L_k\}$, $J$ behaves as a piecewise linear function of $\lambda$, and its partial derivative with respect to $\lambda$ remains constant:
\begin{equation}
 \frac{\partial J}{\partial \lambda} \;=\; \sum_{k=1}^m L_k \;-\; \max_{1\le k\le m}L_k.
\end{equation}
This derivative reflects the difference between the total cost and the maximum individual load. When $\sum_k L_k > \max_k L_k$, increasing $\lambda$ helps to further compress the total cost, whereas reducing $\lambda$ tends to emphasize balancing the worst load. By solving the optimization problem independently under various $\lambda$ values, one can construct a parameterized trade-off curve that implicitly characterizes the Pareto frontier of the dual objectives.

The reconciled objective $J$ not only preserves complete information about both objectives, but also enables a controllable and intuitive relationship between total cost and worst load through a single parameter $\lambda$, thus providing a unified and flexible evaluation criterion for subsequent ACO-based path planning.

To quantify route interference between robots, we treat edges as unordered pairs and define the edge set traversed by tour $P_k$ as
\begin{equation}
E(P_k) = \bigl\{\{i,j\}\in E:\; \{i,j\}\ \text{is traversed by}\ P_k\bigr\},
\end{equation}

For any two tours $P_k$ and $P_l$, we define the (edge level) overlap

\begin{equation}
\begin{aligned}
S(P_k, P_l)
&= \bigl| E(P_k) \cap E(P_l) \bigr| \\
&= \sum_{\{i,j\}\in E} \mathbf{1}_{\{\{i,j\}\in E(P_k)\}}\;\mathbf{1}_{\{\{i,j\}\in E(P_l)\}},\quad k\ne l,
\end{aligned}
\end{equation}

where $\mathbf{1}[\cdot]$ denotes the indicator. This metric counts common edges (ignoring direction) and captures edge level interference.

Furthermore, we enforce the overlap constraint as a hard condition in the model:

\begin{equation}
S(P_k, P_l) = 0, \quad \forall \; 1 \leq k < l \leq m.
\end{equation}

When a soft constraint is preferred, overlap can be penalized in the objective:

\begin{equation}
J' = \lambda \sum_{k=1}^{m} L_k + (1 - \lambda) \max_{k} L_k 
+ \mu \sum_{1 \leq k < l \leq m} S(P_k, P_l),
\end{equation}
where $\mu \ge 0$ controls the strength of overlap penalization.

\subsection{Backbone Based Solution Skeleton}

To establish a robust initial solution, we leverage a minimum cost spanning skeleton as the foundation for constructing a high-quality solution backbone (instantiated in this work via an MST). This skeleton captures essential connectivity with minimal total edge cost, providing an efficient structural prior for subsequent optimization. First, we solve for the skeleton using Kruskal’s algorithm, yielding a low-redundancy spanning tree that connects all nodes with minimal cost: 
\begin{equation}
 T^\ast = \arg\min_{\substack{T\subseteq E,\\T\ \mathrm{connects}}}
 \sum_{(i,j)\in T}d_{ij},\quad |T|=|V|-1.
\end{equation}
This structure defines a low-redundancy, full-coverage backbone that we use to prime optimization. Under the metric assumption, we employ Christofides’ construction (3/2 approximation for the TSP): let $U$ be the set of odd degree vertices of $T^\ast$ and compute a minimum weight perfect matching $M$ on $U$. Then we form the Eulerian multigraph $H=(V, E(T^\ast)\cup M)$, compute an Euler tour in $H$, and obtain a Hamiltonian tour $P_0$ by shortcutting repeated vertices; the triangle inequality ensures non-increasing length during shortcutting. This provides a robust initialization for subsequent local refinements. To make this explicit, we introduce a structural backbone based on the minimum weight perfect matching: 
\begin{equation}
 M = \arg\min_{\substack{M \subseteq E, \\ M \ \mathrm{perfect\ matching\ on}\ U}}
 \sum_{(i,j)\in M} d_{ij}.
\end{equation}
The resulting Eulerian multigraph serves as a backbone skeleton that captures essential connectivity with low cost. 
Note that this construction is not the final solution, but rather a structural initialization. 
In the following stage, this skeleton is used to seed the Ant Colony Optimization (ACO), 
which further refines the tour through probabilistic exploration and reinforcement.

\subsection{ACO with Structural Prior Pheromone Biasing}
To guide probabilistic construction, we introduce a structural prior bias. This amplifies backbone edges within the ACO framework, steering ants toward promising regions while preserving exploration. Let ant $k$ be located at node $v_i$ at iteration $t$. Define the feasible next node set $\mathcal{A}^k_i(t)$ as the unvisited vertices in $V_k$ (or the admissible candidates under constraints). The transition probability is
\begin{equation}
p^k_{ij}(t)
= \frac{\bigl[\tau_{ij}(t)\bigr]^{\alpha}\,\bigl[\eta_{ij}\bigr]^{\beta}\,\bigl[\psi_{ij}\bigr]^{\gamma}}
 {\sum_{\ell\in \mathcal{A}^k_i(t)}\bigl[\tau_{i\ell}(t)\bigr]^{\alpha}\,\bigl[\eta_{i\ell}\bigr]^{\beta}\,\bigl[\psi_{i\ell}\bigr]^{\gamma}},\quad j\in \mathcal{A}^k_i(t),
\end{equation}
where $\tau_{ij}(t)\ge 0$ is the pheromone intensity on edge $(i,j)$, $\eta_{ij}=1/d_{ij}$ is the visibility term with $d_{ij}>0$ for $i\ne j$, and $\psi_{ij}$ is the structural prior bias induced by the backbone:
\begin{equation}
 \psi_{ij} =
\begin{cases}
\omega, & (i,j)\in T^\ast,\\
1, & \text{otherwise},
\end{cases}
\end{equation}
with $\omega\ge 1$. This increases the likelihood of selecting backbone edges without forbidding alternative choices. Unless otherwise noted, the exponents $\alpha,\beta,\gamma>0$ control the relative importance of pheromone ($\alpha$), visibility ($\beta$), and structural prior bias ($\gamma$). 

\subsection{Pheromone Update with Persistent Backbone Influence}
An effective pheromone update is crucial for balancing exploration and exploitation. In SINE, edges aligned with the structural prior receive additional reinforcement while evaporation prevents stagnation. Let $B_k\subseteq E$ denote the set of backbone edges restricted to $V_k$, and let $\kappa\ge 0$ be the backbone influence. Pheromones evolve as
\begin{equation}
\tau_{ij}(t+1) = (1-\rho)\,\tau_{ij}(t) + \sum_{k=1}^m \Delta\tau^k_{ij}(t),
\end{equation}
where $\tau_{ij}(t)$ is the pheromone intensity on edge $(i,j)$ at iteration $t$, $\rho \in (0,1]$ is the evaporation rate, $m$ is the number of ants, and $\Delta\tau^k_{ij}(t)$ is the deposit from ant $k$.

The pheromone deposition is defined as:
\begin{equation}
\Delta\tau^k_{ij}(t) = 
\begin{cases}
\dfrac{Q}{L_k(t)}\left(1 + \kappa\,\mathbf{1}_{\{(i,j)\in B_k\}}\right), & \text{if } (i,j)\in P_k(t), \\[6pt]
0, & \text{otherwise},
\end{cases}
\end{equation}
where:$Q > 0$ is the overall pheromone scaling factor,$L_k(t)$ is the length of tour $P_k(t)$ constructed by ant $k$,$P_k(t)$ is the sequence of edges chosen by ant $k$,$B_k$ is the set of backbone edges restricted to $V_k$,$\kappa \geq 0$ is the structural prior influence factor in deposition,$\mathbf{1}_{\{\cdot\}}$ is the indicator function.

Evaporation multiplies existing pheromone by $(1-\rho)$ to prevent unlimited accumulation, after which robot deposits are added. Each robot deposits only on edges of its current tour, with base amount $Q/L_k(t)$ rewarding shorter tours; the augmentation factor $1 + \kappa\,\mathbf{1}_{\{(i,j)\in B_k\}}$ biases the search toward backbone edges. Tuning $\kappa$ moves smoothly between standard ACO behavior ($\kappa=0$) and strongly prior guided search . In practice, moderate values  provide a good balance between speed and robustness.

\subsection{Multi Ant Agent Path Decomposition}
\begin{figure}[!ht]
 \centering
 \includegraphics[width=0.8\linewidth]{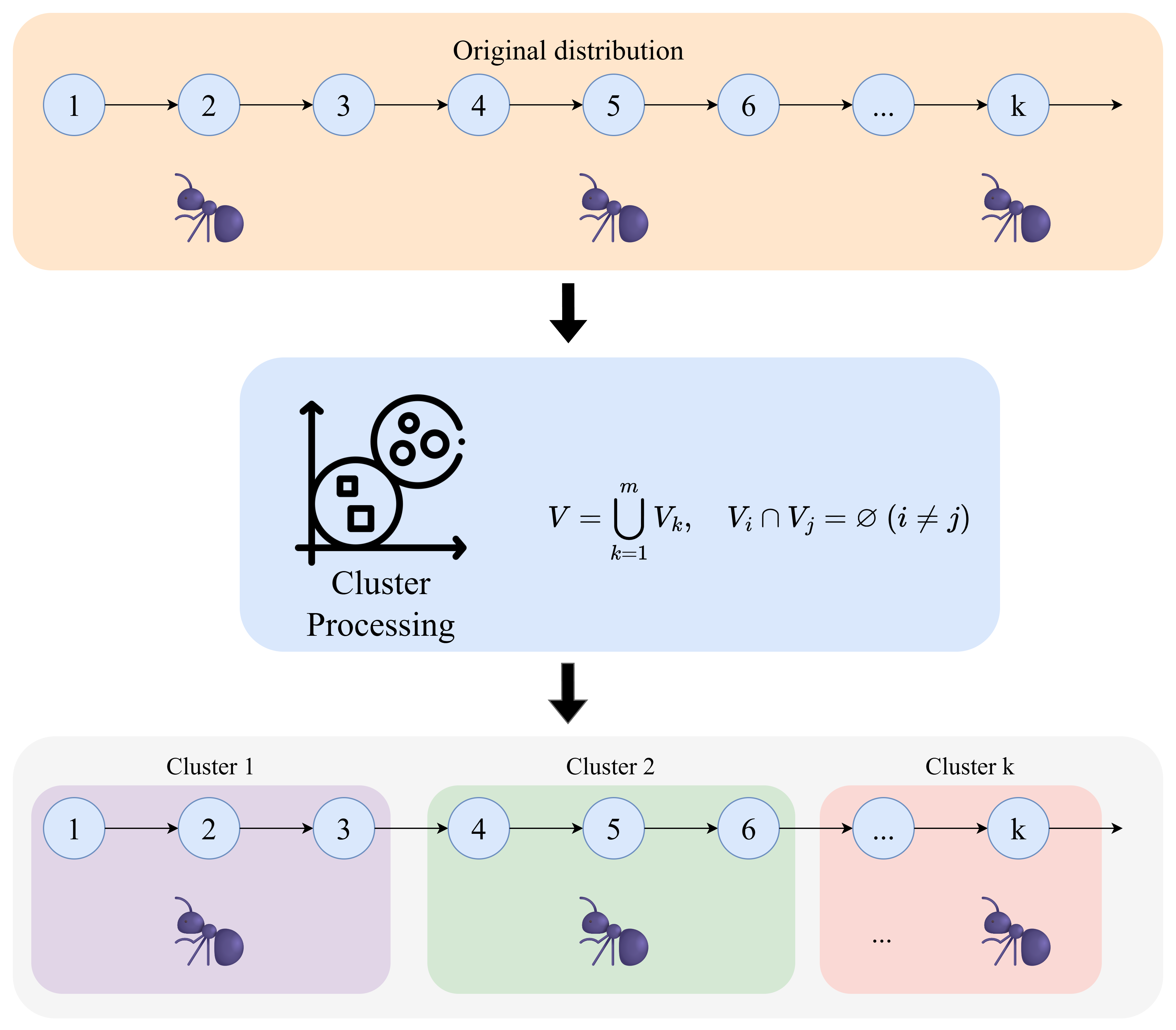}
 \caption{Distributed task allocation for \(k\) ant agents: the full set of \(n\) target nodes \(\{1,2,\dots,n\}\) is partitioned into \(k\) disjoint subsets, each of which is assigned to a different ant for parallel path planning.}
 \label{fig:distributed-allocation}
\end{figure}

 We propose a straightforward yet effective decomposition strategy for multi ant agent path planning. Figure \ref{fig:distributed-allocation} illustrates the task decomposition stage in which the full set of $n$ target points is partitioned into $m$ disjoint subsets for parallel processing (line 9 in Algorithm~\ref{alg:spine-init-construct}). Along the top row, vertices labeled $1,2,3,\dots, K$ represent the complete collection of waypoints to be visited. Beneath these, each ant icon denotes an individual responsible for one of the $k$ clusters: The first ant is assigned to the initial block of points, the second ant to the subsequent block, and this process continues until the $m$-th ant is allocated the final segment. By allocating roughly $\lfloor N/m\rfloor$ or $\lceil N/m\rceil$ nodes per ant, this scheme ensures balanced workload distribution while preserving the underlying Euclidean structure of each subset. Each ant then independently executes an ACO based tour construction (lines 11-22 in Algorithm~\ref{alg:spine-init-construct}), optionally seeded with the backbone derived skeleton restricted to its assigned vertices, thereby enabling concurrent and scalable optimization of the overall multi ant agent routing problem. Group the $n$ target points into $m$ subsets ${V_1, \dots, V_m}$ by angle or clustering algorithm:
\begin{equation}
 V = \bigcup_{k=1}^m V_k,\quad
V_i \cap V_j = \emptyset\quad (i\neq j).
\end{equation}
Each $V_k$ induces an independent Hamiltonian cycle problem. Let $\mathcal{H}(V_k)$ denote the set of all Hamiltonian cycles over $V_k$. We seek $P_k \in \arg\min_{T\in\mathcal{H}(V_k)} L_k(T)$, where $L_k(T)$ is the tour length associated with cycle $T$. To solve each subproblem, we employ an Ant Colony Optimization (ACO) metaheuristic executed in parallel for all ants. The search within $V_k$ can be optionally initialized or biased by restricting the backbone solution to $V_k$, i.e., $T^\ast|_{V_k}$. This decomposition strategy improves scalability while preserving solution quality. We group the $n$ target points into $m$ subsets ${V_1, \dots, V_m}$ by angle or clustering algorithm:
where $L_k(T)$ is the tour length associated with cycle $T$. To solve each subproblem, we employ an Ant Colony Optimization (ACO) metaheuristic executed in parallel for all ants. The search within $V_k$ can be optionally initialized or biased by restricting the backbone solution to $V_k$, i.e., $T^\ast|_{V_k}$. This decomposition strategy improves scalability while preserving solution quality.

\begin{figure}[!ht]
 \centering
 \includegraphics[width=0.25\linewidth]{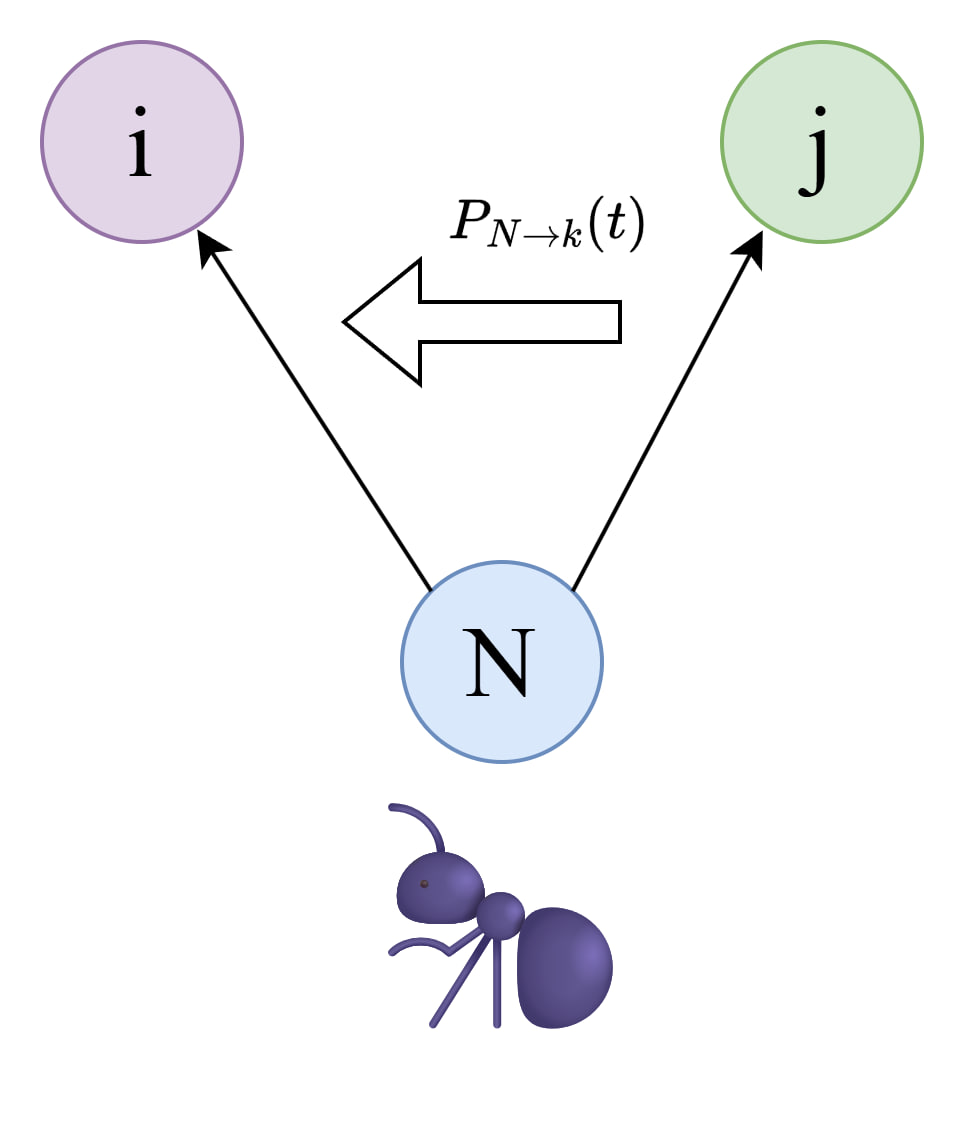}
\caption{Ant transition decision at node \(N\): the probability \(P_{N\to k}(t)\) of moving to candidate vertex \(k\in\{i,j\}\) is computed by normalizing the product of pheromone intensity \(\tau_{Nk}(t)\), heuristic desirability \(\eta_{Nk}=1/d_{Nk}\), and structural-prior bias \(\psi_{Nk}\) over the two options.}

 \label{fig:transition-prob}
\end{figure}

Figure~\ref{fig:transition-prob} illustrates the local decision-making process of an ant located at node $N$ when selecting its next move from two candidate vertices $i$ and $j$ (lines 15-16 in Algorithm~\ref{alg:spine-init-construct}). Each directed arrow emanating from $N$ symbolizes a potential transition whose probability is determined by the relative strength of the pheromone trails $\tau_{Nk}(t)$, the heuristic desirability $\eta_{Nk}=1/d_{Nk}$, and the structural bias $\psi_{Nk}$ that increases the appeal of edges belonging to the backbone. By normalizing the product of these three factors over the available options, the ant stochastically chooses its successor vertex in a manner that balances exploitation of high-quality, tree-guided connections against exploration of alternative routes. This mechanism ensures that early searches are guided by the low-redundancy skeleton provided by the backbone, while still preserving the adaptive, pheromone-driven refinement that characterizes the ACO metaheuristic.

\subsection{SINE Overall Algorithm}

\label{sec:spine}

The SINE algorithm operates through two major phases, presented in Algorithms~\ref{alg:spine-init-construct} and~\ref{alg:spine-update}. In the first phase (Algorithm~\ref{alg:spine-init-construct}), the algorithm begins by computing a structural backbone (instantiated here as an MST) and initializing pheromone values and prior weights. It enters the iterative solution construction process, where each robot constructs its tour over assigned task subsets. This phase combines initialization and iterative optimization to produce feasible solutions guided by persistent structural information. In the solution construction phase, each robot builds its tour based on probabilistic decisions guided by pheromone trails, heuristic desirability, and backbone derived biases in algorithm~\ref{alg:spine-init-construct}. After tours completed, pheromone trails are updated according to the quality of solutions found. This phase incorporates both pheromone evaporation and reinforcement, with additional weight given to edges present in the structural backbone in Algorithm~\ref{alg:spine-update}. Optionally, the algorithm also evaluates the combined objective function and records the best solution found.

\begin{algorithm}[ht]
\caption{SINE Initialization and Solution Construction Phase}
\label{alg:spine-init-construct}
\begin{algorithmic}[1]
\Procedure{SINE\_InitAndConstruct}{$G, m, T_{\max}$}
\Comment{\textbf{MST Initialization}}
\State Compute global MST $T^*$ on $G$
\State Initialize pheromone: $\tau_{ij} \gets \tau_0 > 0$ for all $(i,j) \in E$
\State Define structural bias: $\psi_{ij} \gets \psi$ if $(i,j) \in T^*$, else $1$

\Comment{\textbf{Iterative Solution Construction}}
\For{$t \gets 1$ to $T_{\max}$}
 \For{each ant $k=1 \ldots m$ \textbf{in parallel}}
 \State Assign subset $V_k$ (e.g., angle-based or clustering)
 \State $P_k \gets$ empty tour, pick start vertex $v \in V_k$
 \While{$|P_k| < |V_k|$}
 \State Let $i$ be current vertex
 \State Compute transition probabilities for each $j \in V_k \setminus P_k$:
 \[
 p_{ij} \propto (\tau_{ij})^\alpha \cdot (1/d_{ij})^\beta \cdot (\psi_{ij})^\gamma
 \]
 \State Select next vertex $j$ via roulette-wheel
 \State Append $j$ to $P_k$, move to $j$
 \EndWhile
 \State Close tour: append start vertex to $P_k$
 \State Compute length: $L_k = \sum_{(i,j) \in P_k} d_{ij}$
 \EndFor
\EndFor
\EndProcedure
\end{algorithmic}
\end{algorithm}

\begin{algorithm}[ht]
\caption{SINE Pheromone Update Phase}
\label{alg:spine-update}
\begin{algorithmic}[1]
\Procedure{SINE\_Update}{$m$}
\ForAll{$(i,j) \in E$}
 \State $\tau_{ij} \gets (1 - \rho) \cdot \tau_{ij}$
 \For{$k=1 \ldots m$}
 \If{$(i,j) \in P_k$}
 \State $\Delta \tau^k_{ij} \gets \dfrac{Q}{L_k} \cdot \left(1 + \alpha \cdot \mathbf{1}_{\{(i,j) \in T^*\}}\right)$
 \Else
 \State $\Delta \tau^k_{ij} \gets 0$
 \EndIf
 \State $\tau_{ij} \gets \tau_{ij} + \Delta \tau^k_{ij}$
 \EndFor
\EndFor
\State Optionally compute $J$ and record best tours
\EndProcedure
\end{algorithmic}
\end{algorithm}

\section{Experiment}\label{sec:exp}
\subsection{Experiment Setup}
This study employs a comprehensive experimental framework to evaluate the performance of the proposed SINE algorithm in multi-robot path planning. To ensure a fair and meaningful comparison, all baseline algorithms were extensively fine-tuned using a grid search over key hyperparameters, and their best-performing configurations were selected for evaluation. This tuning process was carried out independently for each algorithm to ensure that they operate under conditions that maximize their respective performance. The experimental design adheres to rigorous scientific principles to ensure the statistical validity and reproducibility of the results. The comparative evaluation includes the above algorithms representing different optimization paradigms:

\begin{itemize}
 \item \textbf{SINE}: The proposed Structural Prior Induced Exploration method
 \item \textbf{ ACO}: Classical Ant Colony Optimization with standard parameter settings
 \item \textbf{AR-ACO}\citep{liu2025high}: Adaptive Rank-based Ant Colony Optimization
 \item \textbf{DL-ACO}\citep{yang2018new}: Deep Learning enhanced Ant Colony Optimization
 \item \textbf{IE-ACO} \citep{li2025intelligently}: Information Entropy enhanced Ant Colony Optimization
 \item \textbf{Smooth ACO}\citep{zhu2011new}: Smoothed pheromone update Ant Colony Optimization
\end{itemize}
\subsection{Dataset Configuration}

The experimental dataset we use, the Traveling Salesman Problem Library, comprises systematically designed test instances that comprehensively cover the problem space of multi-robot path planning scenarios. The dataset characteristics are summarized in Table \ref{tab:dataset_config}. All algorithms are configured with standardized parameters to ensure fair comparison, as detailed in Table \ref{tab:algorithm_params}. The test instances are generated using a stratified sampling approach to ensure comprehensive coverage of problem complexities:

\begin{table}[htbp]
\centering
\caption{Dataset Configuration and Problem Instance Distribution}
\label{tab:dataset_config}
\begin{tabular}{lcc}
\toprule
\textbf{Parameter} & \textbf{Range} &  \\
\midrule
Number of Cities & 50 -- 1,577  \\
Number of Robots & 2 , 6, 8  \\
Problem Instances per Configuration & 8  \\
Total Experimental Runs & 512  \\
\bottomrule
\end{tabular}
\end{table}

\textbf{Small-scale Problems (51--200 cities):} These instances simulate warehouse automation and indoor robotics scenarios where precise coordination is critical. City locations follow clustered distributions mimicking real-world facility layouts.

\textbf{Medium-scale Problems (201--800 cities):} Representative of urban delivery systems and surveillance networks, these instances incorporate both clustered and random geographic distributions to test algorithmic robustness across diverse spatial configurations.

\textbf{Large-scale Problems (801--1,577 cities):} Designed to evaluate scalability for complex logistics operations and large area monitoring systems, these instances challenge algorithms with high-dimensional optimization spaces and increased computational complexity.

\textbf{ Antarctica Dataset:} To complement these synthetic benchmarks, we use a real-world Antarctica dataset\cite{kim2023multi}. In the simulation, the Antarctic environment was located at $74^{\circ}37.4^{\prime}\mathrm{S},\ 164^{\circ}13.7^{\prime}\mathrm{E}$, and nodes were randomly set nearby. 
The latitude and longitude values of arbitrary nodes were extracted from Google Earth. The distance between nodes was calculated using the Haversine Formula, and the altitude values of nodes and edges were obtained using the Google Maps API.

\begin{table}[htbp]
\centering
\caption{Algorithm Parameter Configuration}
\label{tab:algorithm_params}
\begin{tabular}{lcc}
\toprule
\textbf{Parameter} & \textbf{Value}\\
\midrule
Population Size & 50 \\
Maximum Iterations & 1000 \\
Pheromone Evaporation Rate & 0.1 \\
Alpha (pheromone importance) & 1.0 \\
Beta (heuristic importance) & 2.0 \\
Crossover Rate & 0.8 \\
\bottomrule
\end{tabular}
\end{table}

\section{Discussion of Results} \label{sec:Performance Analysis}
This section presents the results of the experimental setup (see Section \ref{sec:exp}) for both the TSPLIB dataset and the real-world Antarctica scenario. To aid the discussion, the results of the TSPLIB were presented using the total tour length (representing global efficiency) and the maximum of individual robot path length (indicating workload balance). The experiments were conducted in a Python environment running on a computer equipped with a 3.50 GHz Intel i9 CPU and 64GB of RAM. All algorithms were run with standardized parameters as detailed in Table~\ref{tab:algorithm_params}.


\subsection{Single Problem Performance Analysis}
In this section, we benchmark our proposed method against other ACO-based algorithms by comparing the performance on each problem instance of the TSPLIB dataset in a multi-robot path planning setting comprising 2, 4, and 8 robots. Additionally, we present the results for the Antarctica scenario, which consists of 50 nodes.

The results for the TSPLIB using a team of 2, 4, and 8 robots are presented in Tables \ref{tab:perf_rc2} to \ref{tab:max_rc8}, respectively. Table \ref{tab:perf_rc2} summarizes the total tour lengths for configurations with two robots across various TSPLIB instances, ranging from small to large scales. SINE consistently achieves the lowest mean tour lengths in all cases, demonstrating clear advantages over the other ACO variants. The summary of the Wilcoxon signed rank test indicates that SINE outperformed all 21 problem instances in the TSPLIB using a team of 2 robots. This edge comes from the backbone's capacity to start with high-quality Eulerian skeletons, which removes unnecessary backtracks early on and steers ants toward Hamiltonian circuits with less deviation. Table \ref{tab:max_rc2} presents the results in terms of maximum path length, which represents the workload distribution among the robots; a high maximum path length indicates that one robot is heavily loaded while others are not well engaged. The table shows that SINE presents the most balanced load distribution (18 out of 21), followed by DL-ACO (3 out of 21) and AR-ACO (1 out of 21). This shows that SINE not only minimizes the total tour lengths but also results in a balanced task distribution among the robots.

\begin{table*}[htbp]
\centering
\caption{Performance Comparison of the Total Path Length (mean~$\pm$~std) for the TSPLIB using a team of 2 Robots. Triangle-up (\triup) = significantly better; diamond (\diamondfill) = statistically equal; triangle-down (\tridown) = significantly worse, according to Wilcoxon signed-rank tests versus the best algorithm in each row.  Best mean values are in \textbf{bold}.}
\label{tab:perf_rc2}
\scriptsize
\resizebox{\textwidth}{!}{
\begin{tabular}{c|cccccc}
\toprule
\textbf{ID} & \textbf{SINE} & \textbf{ACO} & \textbf{AR-ACO} & \textbf{DL-ACO} & \textbf{IEACO} & \textbf{Smooth ACO} \\
\midrule
C51 & \textbf{4.32E+02 $\pm$(6.96E+00)} \triup & 5.20E+02 $\pm$(1.62E+01) \tridown  & 5.07E+02 $\pm$(1.13E+01) \tridown  & 4.80E+02 $\pm$(2.35E+01) \tridown  & 5.55E+02 $\pm$(9.39E+00) \tridown  & 5.15E+02 $\pm$(2.25E+01) \tridown  \\
C52 & \textbf{7.12E+03 $\pm$(1.77E+02)} \triup & 9.84E+03 $\pm$(1.97E+02) \tridown  & 8.55E+03 $\pm$(1.71E+02) \tridown  & 7.91E+03 $\pm$(3.79E+02) \tridown  & 1.04E+04 $\pm$(4.83E+02) \tridown  & 9.64E+03 $\pm$(3.96E+02) \tridown  \\
C70 & \textbf{6.52E+02 $\pm$(8.15E+00)} \triup & 8.44E+02 $\pm$(3.93E+01) \tridown  & 7.24E+02 $\pm$(3.55E+01) \tridown  & 7.76E+02 $\pm$(1.78E+01) \tridown  & 9.46E+02 $\pm$(1.91E+01) \tridown  & 8.16E+02 $\pm$(2.37E+01) \tridown  \\
C76 & \textbf{1.22E+05 $\pm$(1.30E+03)} \triup & 1.60E+05 $\pm$(5.21E+03) \tridown  & 1.27E+05 $\pm$(5.81E+03) \tridown  & 1.27E+05 $\pm$(4.83E+03) \tridown  & 1.89E+05 $\pm$(9.00E+03) \tridown  & 1.31E+05 $\pm$(5.84E+03) \tridown  \\
C99 & \textbf{1.26E+03 $\pm$(5.76E+01)} \triup & 1.64E+03 $\pm$(6.33E+01) \tridown  & 1.40E+03 $\pm$(2.50E+01) \tridown  & 1.45E+03 $\pm$(2.40E+01) \tridown  & 2.24E+03 $\pm$(4.81E+01) \tridown  & 1.50E+03 $\pm$(5.02E+01) \tridown  \\
C198 & \textbf{1.48E+04 $\pm$(6.35E+02)} \triup & 2.42E+04 $\pm$(5.51E+02) \tridown  & 1.64E+04 $\pm$(5.80E+02) \tridown  & 1.79E+04 $\pm$(4.10E+02) \tridown  & 3.69E+04 $\pm$(1.15E+03)  \tridown & 1.78E+04 $\pm$(5.24E+02) \tridown  \\
C318 & \textbf{4.55E+04 $\pm$(1.87E+03)} \triup & 8.21E+04 $\pm$(3.68E+03) \tridown  & 5.05E+04 $\pm$(1.44E+03)  \tridown & 5.22E+04 $\pm$(1.49E+03) \tridown  & 1.37E+05 $\pm$(4.01E+03) \tridown  & 5.80E+04 $\pm$(2.09E+03) \tridown  \\
C417 & \textbf{1.43E+04 $\pm$(4.45E+02)} \triup & 2.19E+04 $\pm$(3.15E+02) \tridown  & 1.59E+04 $\pm$(4.22E+02) \tridown  & 1.73E+04 $\pm$(7.71E+02) \tridown  & 8.36E+04 $\pm$(1.56E+03) \tridown  & 1.92E+04 $\pm$(5.65E+02) \tridown  \\
C442 & \textbf{5.76E+04 $\pm$(1.57E+03)} \triup & 1.14E+05 $\pm$(3.05E+03) \tridown  & 6.39E+04 $\pm$(2.75E+03) \tridown  & 6.76E+04 $\pm$(2.63E+03) \tridown  & 1.79E+05 $\pm$(3.58E+03) \tridown  & 6.18E+04 $\pm$(2.00E+03) \tridown  \\
C574 & \textbf{4.43E+04 $\pm$(2.09E+03)} \triup & 8.22E+04 $\pm$(3.18E+03) \tridown  & 5.00E+04 $\pm$(1.39E+03) \tridown  & 5.18E+04 $\pm$(1.45E+03) \tridown  & 1.58E+05 $\pm$(2.99E+03) \tridown  & 5.76E+04 $\pm$(2.50E+03) \tridown  \\
C783 &\textbf{ 1.03E+04 $\pm$(3.45E+02)} \triup & 1.95E+04 $\pm$(5.96E+02) \tridown  & 1.16E+04 $\pm$(3.93E+02) \tridown  & 1.21E+04 $\pm$(5.83E+02) \tridown  & 4.31E+04 $\pm$(1.84E+03) \tridown  & 1.32E+04 $\pm$(6.09E+02) \tridown  \\
C1002 &\textbf{ 3.02E+05 $\pm$(1.25E+04)} \triup & 6.45E+05 $\pm$(8.58E+03) \tridown  & 3.56E+05 $\pm$(1.64E+04) \tridown  & 3.82E+05 $\pm$(7.36E+03) \tridown  & 1.48E+06 $\pm$(4.05E+04) \tridown  & 4.01E+05 $\pm$(7.61E+03) \tridown  \\
C1060 & \textbf{2.73E+05 $\pm$(1.35E+04)} \triup & 5.70E+05 $\pm$(1.56E+04) \tridown  & 3.15E+05 $\pm$(1.24E+04) \tridown  & 3.38E+05 $\pm$(9.16E+03)  \tridown & 1.50E+06 $\pm$(6.83E+04)  \tridown & 3.75E+05 $\pm$(1.47E+04) \tridown  \\
C1084 &\textbf{ 2.89E+05 $\pm$(1.31E+04)} \triup & 6.01E+05 $\pm$(6.67E+03) \tridown  & 3.40E+05 $\pm$(1.14E+04)  \tridown & 3.49E+05 $\pm$(9.78E+03)  \tridown & 1.89E+06 $\pm$(6.67E+04)  \tridown & 3.67E+05 $\pm$(1.72E+04)  \tridown  \\
C1173 & \textbf{6.48E+04 $\pm$(2.35E+03)} \triup & 1.54E+05 $\pm$(7.27E+03)  \tridown & 8.02E+04 $\pm$(2.04E+03)  \tridown & 8.30E+04 $\pm$(3.66E+03)  \tridown & 3.37E+05 $\pm$(6.00E+03)  \tridown & 8.52E+04 $\pm$(2.34E+03)  \tridown  \\
C1291 & \textbf{6.00E+04 $\pm$(2.44E+03)} \triup & 1.30E+05 $\pm$(1.39E+03)  \tridown & 6.41E+04 $\pm$(1.61E+03)  \tridown & 7.26E+04 $\pm$(2.99E+03)  \tridown & 3.94E+05 $\pm$(1.40E+04)  \tridown & 6.55E+04 $\pm$(2.73E+03)  \tridown \\
C1304 & \textbf{2.96E+05 $\pm$(1.41E+04)} \triup & 6.92E+05 $\pm$(2.08E+04)  \tridown & 3.41E+05 $\pm$(3.78E+03)  \tridown & 3.73E+05 $\pm$(1.29E+04)  \tridown & 2.22E+06 $\pm$(5.07E+04)  \tridown & 4.02E+05 $\pm$(1.10E+04) \tridown \\
C1400 & \textbf{2.49E+04 $\pm$(9.55E+02)} \triup & 4.05E+04 $\pm$(8.58E+02)  \tridown & 2.71E+04 $\pm$(4.67E+02)  \tridown & 3.03E+04 $\pm$(7.24E+02)  \tridown & 3.13E+05 $\pm$(6.03E+03)  \tridown & 3.13E+04 $\pm$(4.75E+02) \tridown \\
C1432 & \textbf{1.83E+05 $\pm$(8.58E+03)} \triup & 4.17E+05 $\pm$(1.39E+04)  \tridown & 2.25E+05 $\pm$(5.80E+03)  \tridown & 2.33E+05 $\pm$(9.33E+03)  \tridown & 9.61E+05 $\pm$(1.36E+04)  \tridown & 2.20E+05 $\pm$(9.48E+03) \tridown \\
C1577 & \textbf{2.67E+04 $\pm$(9.30E+02)} \triup & 5.45E+04 $\pm$(1.92E+03)  \tridown & 3.09E+04 $\pm$(1.31E+03)  \tridown & 3.36E+04 $\pm$(1.55E+03)  \tridown & 2.85E+05 $\pm$(5.77E+03)  \tridown & 3.32E+04 $\pm$(1.08E+03) \tridown \\
\bottomrule
\triup/\tridown/\diamondfill & 21/0/0 & 0/21/0 & 0/21/0 & 0/21/0 & 0/21/0 & 0/21/0 \\
\bottomrule
\end{tabular}
}
\end{table*}

\begin{table*}[htbp]
\centering
\caption{Performance Comparison of the Maximum Path Length (mean~$\pm$~std) for the TSPLIB using a team of 2 Robots. Triangle-up (\triup) = significantly better; diamond (\diamondfill) = statistically equal; triangle-down (\tridown) = significantly worse, according to Wilcoxon signed-rank tests versus the best algorithm in each row.  Best mean values are in \textbf{bold}.}
\label{tab:max_rc2}
\scriptsize
\resizebox{\textwidth}{!}{
\begin{tabular}{c|cccccc}
\toprule
\textbf{ID} & \textbf{SINE} & \textbf{ACO} & \textbf{AR-ACO} & \textbf{DL-ACO} & \textbf{IEACO} & \textbf{Smooth ACO} \\
\midrule
C51   & 2.59E+02 $\pm$(2.53E+01) \tridown & 2.87E+02 $\pm$(3.71E+01) \tridown & 2.77E+02 $\pm$(3.18E+01) \tridown & \textbf{2.52E+02 $\pm$(1.55E+01)}\triup & 3.06E+02 $\pm$(3.57E+01) \tridown & 2.73E+02 $\pm$(5.56E-01) \tridown \\
C52   & 5.80E+03 $\pm$(4.55E+02) \tridown & 6.28E+03 $\pm$(8.61E+02) \tridown & 5.50E+03 $\pm$(4.11E+02) \tridown & \textbf{5.17E+03 $\pm$(1.27E+01)}\triup & 6.52E+03 $\pm$(3.41E+02) \tridown & 5.59E+03 $\pm$(8.19E+02) \tridown \\
C70   & \textbf{3.93E+02 $\pm$(3.00E+00)}\triup & 4.43E+02 $\pm$(9.44E+01) \tridown & 4.52E+02 $\pm$(1.19E+02) \tridown & 4.32E+02 $\pm$(7.95E+01) \tridown & 4.79E+02 $\pm$(3.29E+00) \tridown & 4.59E+02 $\pm$(2.55E+01) \tridown \\
C76   & \textbf{4.34E+04 $\pm$(3.75E+04)}\triup & 6.15E+04 $\pm$(5.31E+04) \tridown & 5.30E+04 $\pm$(4.58E+04) \tridown & 5.05E+04 $\pm$(4.38E+04) \tridown & 6.78E+04 $\pm$(5.85E+04) \tridown & 5.39E+04 $\pm$(4.68E+04) \tridown \\
C99   & \textbf{7.35E+02 $\pm$(3.42E+01)}\triup & 8.40E+02 $\pm$(1.04E+02) \tridown & 8.13E+02 $\pm$(7.57E+01) \tridown & 8.59E+02 $\pm$(1.21E+01) \tridown & 1.07E+03 $\pm$(1.22E+02) \tridown & 8.24E+02 $\pm$(3.75E+01) \tridown \\
C198   & 1.12E+04 $\pm$(1.93E+03) \tridown & 1.30E+04 $\pm$(3.49E+03) \tridown & \textbf{1.04E+04 $\pm$(3.53E+02)}\triup & 1.10E+04 $\pm$(1.68E+02) \tridown & 1.78E+04 $\pm$(2.43E+03) \tridown & 1.10E+04 $\pm$(2.48E+01) \tridown \\
C318   & \textbf{2.61E+04 $\pm$(2.41E+03)}\triup & 4.46E+04 $\pm$(1.13E+04) \tridown & 3.25E+04 $\pm$(5.07E+03) \tridown & 3.32E+04 $\pm$(2.51E+03) \tridown & 7.16E+04 $\pm$(1.59E+04) \tridown & 3.64E+04 $\pm$(2.59E+03) \tridown \\
C417   & \textbf{8.46E+03 $\pm$(1.44E+01)}\triup & 1.18E+04 $\pm$(2.55E+03) \tridown & 8.47E+03 $\pm$(3.34E+02) \tridown & 8.51E+03 $\pm$(1.67E+03) \tridown & 3.65E+04 $\pm$(7.92E+03) \tridown & 1.06E+04 $\pm$(2.60E+03) \tridown \\
C442   & \textbf{2.84E+04 $\pm$(6.59E+02)}\triup & 5.58E+04 $\pm$(1.70E+04) \tridown & 4.08E+04 $\pm$(5.05E+03) \tridown & 3.87E+04 $\pm$(2.40E+03) \tridown & 8.83E+04 $\pm$(2.29E+04) \tridown & 3.93E+04 $\pm$(5.68E+03) \tridown \\
C574   & \textbf{2.24E+04 $\pm$(1.38E+03)}\triup & 4.25E+04 $\pm$(1.15E+04) \tridown & 3.00E+04 $\pm$(1.09E+03) \tridown & 2.96E+04 $\pm$(3.82E+01) \tridown & 6.92E+04 $\pm$(2.17E+04) \tridown & 3.56E+04 $\pm$(2.60E+01) \tridown \\
C783   & \textbf{5.23E+03 $\pm$(1.43E+02)}\triup & 9.57E+03 $\pm$(3.34E+03) \tridown & 6.74E+03 $\pm$(4.84E+02) \tridown & 7.13E+03 $\pm$(1.18E+02) \tridown & 1.94E+04 $\pm$(6.46E+03) \tridown & 7.70E+03 $\pm$(1.77E+02) \tridown \\
C1002   & \textbf{1.52E+05 $\pm$(1.10E+04)}\triup & 3.13E+05 $\pm$(1.01E+05) \tridown & 2.16E+05 $\pm$(1.31E+03) \tridown & 2.35E+05 $\pm$(1.86E+04) \tridown & 6.83E+05 $\pm$(2.83E+05) \tridown & 2.50E+05 $\pm$(1.49E+04) \tridown \\
C1060   & \textbf{1.34E+05 $\pm$(1.12E+04)}\triup & 2.80E+05 $\pm$(1.08E+05) \tridown & 1.98E+05 $\pm$(1.98E+04) \tridown & 2.00E+05 $\pm$(1.18E+04) \tridown & 6.61E+05 $\pm$(2.42E+05) \tridown & 2.37E+05 $\pm$(1.41E+04) \tridown \\
C1084   & \textbf{1.58E+05 $\pm$(3.65E+04)}\triup & 3.35E+05 $\pm$(1.26E+05) \tridown & 2.23E+05 $\pm$(1.37E+04) \tridown & 2.34E+05 $\pm$(1.01E+03) \tridown & 7.89E+05 $\pm$(4.17E+05) \tridown & 2.63E+05 $\pm$(1.38E+04) \tridown \\
C1173   & \textbf{3.22E+04 $\pm$(5.86E+02)}\triup & 6.80E+04 $\pm$(2.61E+04) \tridown & 4.80E+04 $\pm$(2.10E+02) \tridown & 5.05E+04 $\pm$(1.58E+02) \tridown & 1.49E+05 $\pm$(7.37E+04) \tridown & 5.37E+04 $\pm$(3.64E+03) \tridown \\
C1291   & \textbf{3.48E+04 $\pm$(3.48E+02)}\triup & 7.79E+04 $\pm$(6.66E+02) \tridown & 3.53E+04 $\pm$(5.68E+02) \tridown & 4.10E+04 $\pm$(4.92E+02) \tridown & 2.15E+05 $\pm$(6.44E+02) \tridown & 3.80E+04 $\pm$(6.18E+02) \tridown \\
C1304   & \textbf{1.48E+05 $\pm$(1.48E+03)}\triup & 4.28E+05 $\pm$(2.43E+03) \tridown & 2.05E+05 $\pm$(2.04E+03) \tridown & 2.15E+05 $\pm$(2.35E+03) \tridown & 1.22E+06 $\pm$(2.32E+03) \tridown & 2.39E+05 $\pm$(2.68E+03) \tridown \\
C1400   & \textbf{1.27E+04 $\pm$(1.27E+02)}\triup & 2.86E+04 $\pm$(1.62E+02) \tridown & 1.39E+04 $\pm$(2.10E+02) \tridown & 1.99E+04 $\pm$(2.26E+02) \tridown & 1.58E+05 $\pm$(2.13E+02) \tridown & 2.14E+04 $\pm$(2.47E+02) \tridown \\
C1432   & \textbf{9.49E+04 $\pm$(9.49E+02)}\triup & 2.33E+05 $\pm$(1.79E+03) \tridown & 1.32E+05 $\pm$(1.55E+03) \tridown & 1.41E+05 $\pm$(1.78E+03) \tridown & 5.49E+05 $\pm$(1.16E+03) \tridown & 1.29E+05 $\pm$(1.15E+03) \tridown \\
C1577   & \textbf{1.43E+04 $\pm$(1.43E+02)}\triup & 3.50E+04 $\pm$(2.65E+02) \tridown & 1.71E+04 $\pm$(1.86E+02) \tridown & 2.00E+04 $\pm$(2.81E+02) \tridown & 1.44E+05 $\pm$(2.53E+02) \tridown & 1.82E+04 $\pm$(2.76E+02) \tridown \\
\bottomrule
\triup/\tridown/\diamondfill & 18/3/0 & 0/21/0 & 1/20/0 & 2/19/0 & 0/21/0 & 0/21/0 \\
\bottomrule
\end{tabular}
}
\end{table*}

As the team size grows to four and eight robots, coordination becomes more intricate; SINE scales in performance, showing smaller total path length and maximum path relative to the baselines. Table \ref{tab:perf_rc4} shows the result of the optimized total path length for the team of four robots obtained by each algorithm. The results show that our proposed SINE algorithm achieved the best performance compared to other variants of ACO in all instances of the TSPLIB. This similar performance is obtained from the team of eight robots as shown in Table \ref{tab:perf_rc8}, where SINE also demonstrates a superior performance across all the TSPLIB instances. In terms of the load distribution, the maximum path length for the team of four robots is presented in Table \ref{tab:max_rc4}. The summarized Wilcoxon signed-rank test shows that SINE achieves the lowest maximum path length in 18 out of the 21 problems, while DL-ACO achieves the lowest maximum path length in 3 out of 21 problems. For a team of eight robots, SINE achieves the lowest maximum path length in 17 out of 21 problems, AR-ACO in 2 out of 21 and DL-ACO in 2 out of 21 problems. This shows that the performance of SINE is scalable and leads to balanced exploration; proving that this framework is effective in navigating complex multi-agent environments.

The proposed SINE facilitates a more targeted search process, limiting initial exploration to promising, low-cost connections while allowing adaptive refinement through pheromone updates and load-aware objectives. This approach yields solutions that optimize overall efficiency and promote balanced task distribution among agents, minimizing redundant overlaps and reducing variance in individual workloads. In larger scenarios, SINE delivers stronger outcomes than DL-ACO and IEACO, with enhanced stability reflected in reduced variance compared to traditional ACO approaches. This consistency arises from adaptive pheromone updates that favor paths aligned with the backbone while supporting random exploration. Statistical evaluations highlight SINE's reliability in most scenarios, particularly in randomly distributed setups where baselines tend to underperform due to unfocused search strategies. In rigorous configurations with 8 robots, as shown in Table~\ref{tab:perf_rc8}, SINE handles scaling well, yielding substantial enhancements in expansive instances. In general, SINE yields stable variability and scalability, proving its effectiveness in navigating the rapid expansion of the solution space through guided exploration.

\begin{table*}[htbp]
\centering
\caption{Performance Comparison of the Total Path Length (mean~$\pm$~std) for the TSPLIB using a team of 4 Robots. Triangle-up (\triup) = significantly better; diamond (\diamondfill) = statistically equal; triangle-down (\tridown) = significantly worse, according to Wilcoxon signed-rank tests versus the best algorithm in each row.  Best mean values are in \textbf{bold}.}
\label{tab:perf_rc4}
\scriptsize
\resizebox{\textwidth}{!}{
\begin{tabular}{c|cccccc}
\toprule
\textbf{ID} & \textbf{SINE} & \textbf{ACO} & \textbf{AR-ACO} & \textbf{DL-ACO} & \textbf{IEACO} & \textbf{Smooth ACO} \\
\midrule
C51   & \textbf{4.57E+02 $\pm$(6.62E+00)} \triup & 5.40E+02 $\pm$(2.27E+01)  \tridown & 5.25E+02 $\pm$(5.64E+00)  \tridown & 5.07E+02 $\pm$(1.77E+01)  \tridown & 6.22E+02 $\pm$(8.85E+00)  \tridown & 5.64E+02 $\pm$(2.92E+01) \tridown \\
C52   & \textbf{8.76E+03 $\pm$(4.23E+02)} \triup & 1.19E+04 $\pm$(5.80E+02)  \tridown & 9.73E+03 $\pm$(4.10E+02)  \tridown & 9.93E+03 $\pm$(1.64E+02)  \tridown & 1.18E+04 $\pm$(2.48E+02)  \tridown & 1.06E+04 $\pm$(1.51E+02) \tridown \\
C70   & \textbf{7.18E+02 $\pm$(2.82E+01)} \triup & 8.41E+02 $\pm$(2.34E+01)  \tridown & 7.98E+02 $\pm$(3.25E+01)  \tridown & 8.07E+02 $\pm$(8.23E+00)  \tridown & 1.11E+03 $\pm$(4.07E+01)  \tridown & 8.61E+02 $\pm$(3.97E+01) \tridown \\
C76   & \textbf{1.38E+05 $\pm$(3.42E+03)} \triup & 1.86E+05 $\pm$(3.35E+03)  \tridown & 1.53E+05 $\pm$(1.81E+03)  \tridown & 1.56E+05 $\pm$(1.74E+03)  \tridown & 1.96E+05 $\pm$(7.48E+03)  \tridown & 1.60E+05 $\pm$(7.54E+03) \tridown \\
C99   & \textbf{1.37E+03 $\pm$(5.69E+01)} \triup & 1.78E+03 $\pm$(6.59E+01)  \tridown & 1.54E+03 $\pm$(8.23E+00)  \tridown & 1.52E+03 $\pm$(5.86E+01)  \tridown & 2.36E+03 $\pm$(7.30E+01)  \tridown & 1.72E+03 $\pm$(1.06E+01) \tridown \\
C198  & \textbf{1.87E+04 $\pm$(2.94E+02)} \triup & 2.85E+04 $\pm$(1.18E+03)  \tridown & 2.08E+04 $\pm$(2.74E+02)  \tridown & 2.12E+04 $\pm$(3.62E+02)  \tridown & 4.04E+04 $\pm$(1.08E+03)  \tridown & 2.26E+04 $\pm$(5.83E+02) \tridown \\
C318  & \textbf{5.48E+04 $\pm$(1.75E+03)} \triup & 9.07E+04 $\pm$(2.89E+03)  \tridown & 5.83E+04 $\pm$(1.59E+03)  \tridown & 5.74E+04 $\pm$(1.96E+03)  \tridown & 1.34E+05 $\pm$(2.90E+03)  \tridown & 6.42E+04 $\pm$(9.83E+02) \tridown \\
C417  & \textbf{1.74E+04 $\pm$(6.00E+02)} \triup & 2.57E+04 $\pm$(8.21E+02)  \tridown & 1.93E+04 $\pm$(5.68E+02)  \tridown & 2.17E+04 $\pm$(6.26E+02)  \tridown & 9.08E+04 $\pm$(4.37E+03)  \tridown & 2.25E+04 $\pm$(5.83E+02) \tridown \\
C442  & \textbf{6.20E+04 $\pm$(2.76E+03)} \triup & 1.20E+05 $\pm$(2.06E+03)  \tridown & 6.82E+04 $\pm$(2.54E+03)  \tridown & 7.31E+04 $\pm$(1.16E+03)  \tridown & 1.89E+05 $\pm$(4.03E+03)  \tridown & 7.04E+04 $\pm$(1.58E+03) \tridown \\
C500  & \textbf{2.00E+03 $\pm$(2.95E+01)} \triup & 2.93E+03 $\pm$(4.43E+01)  \tridown & 2.20E+03 $\pm$(5.93E+01)  \tridown & 2.32E+03 $\pm$(6.24E+01)  \tridown & 6.37E+03 $\pm$(2.52E+02)  \tridown & 2.54E+03 $\pm$(7.08E+01) \tridown \\
C574  & \textbf{5.02E+04 $\pm$(1.64E+03)} \triup & 9.10E+04 $\pm$(3.28E+03)  \tridown & 5.40E+04 $\pm$(2.40E+03)  \tridown & 5.67E+04 $\pm$(2.72E+03)  \tridown & 1.63E+05 $\pm$(2.02E+03)  \tridown & 6.17E+04 $\pm$(1.39E+03) \tridown \\
C783  & \textbf{1.10E+04 $\pm$(2.69E+02)} \triup & 2.01E+04 $\pm$(7.81E+02)  \tridown & 1.23E+04 $\pm$(3.38E+02)  \tridown & 1.27E+04 $\pm$(4.71E+02)  \tridown & 4.43E+04 $\pm$(1.23E+03)  \tridown & 1.40E+04 $\pm$(4.53E+02) \tridown \\
C1002 & \textbf{3.29E+05 $\pm$(1.02E+04)} \triup & 6.67E+05 $\pm$(1.70E+04)  \tridown & 3.83E+05 $\pm$(1.77E+04)  \tridown & 3.97E+05 $\pm$(9.45E+03)  \tridown & 1.54E+06 $\pm$(6.56E+04)  \tridown & 4.43E+05 $\pm$(1.68E+04) \tridown \\
C1060 & \textbf{2.96E+05 $\pm$(1.42E+04)} \triup & 6.03E+05 $\pm$(1.57E+04)  \tridown & 3.48E+05 $\pm$(1.71E+04)  \tridown & 3.61E+05 $\pm$(5.45E+03)  \tridown & 1.55E+06 $\pm$(6.31E+04)  \tridown & 4.06E+05 $\pm$(1.78E+04) \tridown \\
C1084 & \textbf{3.16E+05 $\pm$(1.15E+04)} \triup & 6.09E+05 $\pm$(2.28E+04)  \tridown & 3.48E+05 $\pm$(4.21E+03)  \tridown & 3.74E+05 $\pm$(1.07E+04)  \tridown & 1.83E+06 $\pm$(7.80E+04)  \tridown & 4.05E+05 $\pm$(4.22E+03) \tridown \\
C1173 & \textbf{7.08E+04 $\pm$(2.43E+03)} \triup & 1.57E+05 $\pm$(6.04E+03)  \tridown & 8.31E+04 $\pm$(3.90E+03)  \tridown & 8.79E+04 $\pm$(1.93E+03)  \tridown & 3.28E+05 $\pm$(9.81E+03)  \tridown & 8.72E+04 $\pm$(9.16E+02) \tridown \\
C1291 & \textbf{6.46E+04 $\pm$(2.84E+03)} \triup & 1.34E+05 $\pm$(2.45E+03)  \tridown & 7.23E+04 $\pm$(2.79E+03)  \tridown & 7.53E+04 $\pm$(1.17E+03)  \tridown & 3.93E+05 $\pm$(1.75E+04)  \tridown & 7.39E+04 $\pm$(1.92E+03) \tridown \\
C1304 & \textbf{3.26E+05 $\pm$(1.50E+04)} \triup & 7.24E+05 $\pm$(8.11E+03)  \tridown & 3.73E+05 $\pm$(6.56E+03)  \tridown & 4.16E+05 $\pm$(6.98E+03)  \tridown & 2.16E+06 $\pm$(8.98E+04)  \tridown & 4.25E+05 $\pm$(1.99E+04) \tridown \\
C1400 & \textbf{2.85E+04 $\pm$(4.05E+02)} \triup & 4.44E+04 $\pm$(1.63E+03)  \tridown & 3.16E+04 $\pm$(4.04E+02)  \tridown & 3.16E+04 $\pm$(1.15E+03)  \tridown & 3.16E+05 $\pm$(1.55E+04)  \tridown & 3.51E+04 $\pm$(1.12E+03) \tridown \\
C1432 & \textbf{1.99E+05 $\pm$(5.24E+03)} \triup & 4.34E+05 $\pm$(8.59E+03)  \tridown & 2.40E+05 $\pm$(6.17E+03)  \tridown & 2.49E+05 $\pm$(1.12E+04)  \tridown & 9.84E+05 $\pm$(4.06E+04)  \tridown & 2.37E+05 $\pm$(7.39E+03) \tridown \\
C1577 & \textbf{3.06E+04 $\pm$(1.35E+03)} \triup & 5.92E+04 $\pm$(1.62E+03)  \tridown & 3.47E+04 $\pm$(7.49E+02)  \tridown & 3.92E+04 $\pm$(1.91E+03)  \tridown & 2.84E+05 $\pm$(4.94E+03)  \tridown & 3.79E+04 $\pm$(1.18E+03) \tridown \\
\bottomrule
\triup/\tridown/\diamondfill & 21/0/0 & 0/21/0 & 0/21/0 & 0/21/0 & 0/21/0 & 0/21/0 \\
\bottomrule
\end{tabular}
}
\end{table*}

\begin{table*}[htbp]
\centering
\caption{Performance Comparison of the Maximum Path Length (mean~$\pm$~std) for the TSPLIB using a team of 4 Robots. Triangle-up (\triup) = significantly better; diamond (\diamondfill) = statistically equal; triangle-down (\tridown) = significantly worse, according to Wilcoxon signed-rank tests versus the best algorithm in each row.  Best mean values are in \textbf{bold}.}
\label{tab:max_rc4}
\scriptsize
\resizebox{\textwidth}{!}{
\begin{tabular}{c|cccccc}
\toprule
\textbf{ID} & \textbf{SINE} & \textbf{ACO} & \textbf{AR-ACO} & \textbf{DL-ACO} & \textbf{IEACO} & \textbf{Smooth ACO} \\
\midrule
C51   & 1.49E+02 $\pm$(2.32E+01) \tridown & 1.70E+02 $\pm$(1.27E+01) \tridown & 1.53E+02 $\pm$(2.44E+01) \tridown & \textbf{1.49E+02 $\pm$(1.58E+01)}\triup & 1.88E+02 $\pm$(3.34E+01) \tridown & 1.60E+02 $\pm$(8.38E+00) \tridown \\
C52   & 3.73E+03 $\pm$(6.69E+02) \tridown & 4.35E+03 $\pm$(7.17E+02) \tridown & 3.49E+03 $\pm$(8.11E+02) \tridown & \textbf{3.32E+03 $\pm$(5.21E+02)}\triup & 4.04E+03 $\pm$(5.04E+02) \tridown & 3.96E+03 $\pm$(7.72E+02) \tridown \\
C70   & \textbf{2.34E+02 $\pm$(3.38E+00)}\triup & 2.67E+02 $\pm$(1.21E+01) \tridown & 2.98E+02 $\pm$(6.82E+01) \tridown & 2.66E+02 $\pm$(1.35E+01) \tridown & 3.23E+02 $\pm$(3.37E+01) \tridown & 3.03E+02 $\pm$(1.93E+01) \tridown \\
C76   & \textbf{2.45E+04 $\pm$(2.23E+04)}\triup & 3.45E+04 $\pm$(3.15E+04) \tridown & 2.61E+04 $\pm$(2.39E+04) \tridown & 2.99E+04 $\pm$(2.76E+04) \tridown & 4.03E+04 $\pm$(3.66E+04) \tridown & 2.83E+04 $\pm$(2.57E+04) \tridown \\
C99   & \textbf{4.64E+02 $\pm$(1.92E+01)}\triup & 5.11E+02 $\pm$(5.58E+01) \tridown & 4.99E+02 $\pm$(5.03E+01) \tridown & 4.95E+02 $\pm$(6.56E+01) \tridown & 7.54E+02 $\pm$(9.99E+01) \tridown & 5.00E+02 $\pm$(6.04E+00) \tridown \\
C198   & \textbf{8.41E+03 $\pm$(2.99E+02)}\triup & 9.70E+03 $\pm$(1.41E+03) \tridown & 8.53E+03 $\pm$(3.88E+02) \tridown & 8.69E+03 $\pm$(3.08E+02) \tridown & 1.29E+04 $\pm$(3.03E+02) \tridown & 9.01E+03 $\pm$(6.09E+02) \tridown \\
C318   & \textbf{1.50E+04 $\pm$(1.53E+03)}\triup & 2.65E+04 $\pm$(7.47E+03) \tridown & 2.19E+04 $\pm$(1.56E+03) \tridown & 1.98E+04 $\pm$(1.93E+03) \tridown & 4.22E+04 $\pm$(3.80E+03) \tridown & 2.19E+04 $\pm$(3.41E+02) \tridown \\
C417   & 6.61E+03 $\pm$(3.60E+02) \tridown & 6.91E+03 $\pm$(1.05E+03) \tridown & 6.11E+03 $\pm$(4.14E+01) \tridown & \textbf{5.81E+03 $\pm$(6.36E+02)}\triup & 2.12E+04 $\pm$(4.56E+03) \tridown & 7.24E+03 $\pm$(1.37E+03) \tridown \\
C442   & \textbf{1.73E+04 $\pm$(5.83E+02)}\triup & 2.98E+04 $\pm$(8.85E+03) \tridown & 2.37E+04 $\pm$(3.76E+03) \tridown & 2.60E+04 $\pm$(3.48E+02) \tridown & 4.94E+04 $\pm$(1.71E+04) \tridown & 2.68E+04 $\pm$(5.41E+03) \tridown \\
C500   & \textbf{5.18E+02 $\pm$(5.18E+00)}\triup & 9.76E+02 $\pm$(7.52E+00) \tridown & 7.97E+02 $\pm$(6.57E+00) \tridown & 7.88E+02 $\pm$(8.39E+00) \tridown & 2.22E+03 $\pm$(8.49E+00) \tridown & 8.25E+02 $\pm$(9.35E+00) \tridown \\
C574   & \textbf{1.23E+04 $\pm$(7.99E+02)}\triup & 2.56E+04 $\pm$(5.64E+03) \tridown & 1.96E+04 $\pm$(1.55E+03) \tridown & 1.82E+04 $\pm$(1.76E+03) \tridown & 3.47E+04 $\pm$(8.60E+03) \tridown & 2.12E+04 $\pm$(5.42E+02) \tridown \\
C783   & \textbf{2.85E+03 $\pm$(1.29E+02)}\triup & 5.35E+03 $\pm$(1.43E+03) \tridown & 4.04E+03 $\pm$(1.14E+02) \tridown & 4.59E+03 $\pm$(3.82E+01) \tridown & 1.04E+04 $\pm$(2.90E+03) \tridown & 4.71E+03 $\pm$(1.87E+02) \tridown \\
C1002   & \textbf{8.15E+04 $\pm$(6.36E+03)}\triup & 1.98E+05 $\pm$(6.72E+04) \tridown & 1.27E+05 $\pm$(1.02E+03) \tridown & 1.41E+05 $\pm$(1.48E+04) \tridown & 3.48E+05 $\pm$(8.03E+04) \tridown & 1.73E+05 $\pm$(1.65E+04) \tridown \\
C1060   & \textbf{7.36E+04 $\pm$(4.88E+03)}\triup & 1.80E+05 $\pm$(6.44E+04) \tridown & 1.38E+05 $\pm$(9.98E+03) \tridown & 1.35E+05 $\pm$(8.23E+03) \tridown & 3.56E+05 $\pm$(1.06E+05) \tridown & 1.60E+05 $\pm$(7.66E+03) \tridown \\
C1084   & \textbf{9.12E+04 $\pm$(9.14E+03)}\triup & 1.92E+05 $\pm$(6.15E+04) \tridown & 1.31E+05 $\pm$(1.21E+04) \tridown & 1.59E+05 $\pm$(1.20E+04) \tridown & 3.75E+05 $\pm$(1.53E+05) \tridown & 1.82E+05 $\pm$(2.49E+04) \tridown \\
C1173   & \textbf{1.81E+04 $\pm$(1.37E+03)}\triup & 3.98E+04 $\pm$(1.32E+04) \tridown & 2.92E+04 $\pm$(2.40E+03) \tridown & 3.26E+04 $\pm$(1.70E+02) \tridown & 7.05E+04 $\pm$(2.37E+04) \tridown & 3.67E+04 $\pm$(1.67E+03) \tridown \\
C1291   & \textbf{1.99E+04 $\pm$(1.99E+02)}\triup & 5.25E+04 $\pm$(3.02E+02) \tridown & 2.59E+04 $\pm$(3.53E+02) \tridown & 2.55E+04 $\pm$(3.81E+02) \tridown & 1.31E+05 $\pm$(3.67E+02) \tridown & 2.49E+04 $\pm$(3.50E+02) \tridown \\
C1304   & \textbf{8.74E+04 $\pm$(8.74E+02)}\triup & 2.47E+05 $\pm$(1.70E+03) \tridown & 1.28E+05 $\pm$(1.41E+03) \tridown & 1.62E+05 $\pm$(1.08E+03) \tridown & 7.19E+05 $\pm$(1.32E+03) \tridown & 1.43E+05 $\pm$(1.32E+03) \tridown \\
C1400   & \textbf{8.42E+03 $\pm$(8.42E+01)}\triup & 2.29E+04 $\pm$(1.66E+02) \tridown & 1.21E+04 $\pm$(1.64E+02) \tridown & 1.15E+04 $\pm$(1.29E+02) \tridown & 8.67E+04 $\pm$(1.26E+02) \tridown & 1.44E+04 $\pm$(1.45E+02) \tridown \\
C1432   & \textbf{5.33E+04 $\pm$(5.33E+02)}\triup & 1.38E+05 $\pm$(7.57E+02) \tridown & 8.79E+04 $\pm$(8.54E+02) \tridown & 8.91E+04 $\pm$(8.17E+02) \tridown & 2.93E+05 $\pm$(7.46E+02) \tridown & 8.61E+04 $\pm$(6.43E+02) \tridown \\
C1577   & \textbf{9.02E+03 $\pm$(9.02E+01)}\triup & 2.43E+04 $\pm$(1.60E+02) \tridown & 9.96E+03 $\pm$(1.57E+02) \tridown & 1.62E+04 $\pm$(1.78E+02) \tridown & 7.91E+04 $\pm$(1.25E+02) \tridown & 1.33E+04 $\pm$(1.58E+02) \tridown \\
\bottomrule
\triup/\tridown/\diamondfill & 18/3/0 & 0/21/0 & 0/21/0 & 3/18/0 & 0/21/0 & 0/21/0 \\
\bottomrule
\end{tabular}
}
\end{table*}

\begin{table*}[htbp]
\centering
\caption{Performance Comparison of the Total Path Length (mean~$\pm$~std) for the TSPLIB using a team of 8 Robots. Triangle-up (\triup) = significantly better; diamond (\diamondfill) = statistically equal; triangle-down (\tridown) = significantly worse, according to Wilcoxon signed-rank tests versus the best algorithm in each row.  Best mean values are in \textbf{bold}.}
\label{tab:perf_rc8}
\scriptsize
\resizebox{\textwidth}{!}{
\begin{tabular}{c|cccccc}
\toprule
\textbf{ID} & \textbf{SINE} & \textbf{ACO} & \textbf{AR-ACO} & \textbf{DL-ACO} & \textbf{IEACO} & \textbf{Smooth ACO} \\
\midrule
C51   & \textbf{5.34E+02 $\pm$(1.82E+01)}\triup & 5.93E+02 $\pm$(1.04E+01) \tridown & 5.95E+02 $\pm$(2.02E+01) \tridown & 6.12E+02 $\pm$(2.15E+01) \tridown & 6.51E+02 $\pm$(2.72E+01) \tridown & 6.63E+02 $\pm$(3.17E+01) \tridown \\
C52   & \textbf{1.18E+04 $\pm$(4.79E+02)}\triup & 1.52E+04 $\pm$(1.85E+02) \tridown & 1.31E+04 $\pm$(4.02E+02) \tridown & 1.37E+04 $\pm$(3.37E+02) \tridown & 1.54E+04 $\pm$(6.40E+02) \tridown & 1.41E+04 $\pm$(2.01E+02) \tridown \\
C70   & \textbf{8.28E+02 $\pm$(1.04E+01)}\triup & 9.87E+02 $\pm$(2.62E+01) \tridown & 9.20E+02 $\pm$(1.42E+01) \tridown & 9.82E+02 $\pm$(2.48E+01) \tridown & 1.16E+03 $\pm$(5.72E+01) \tridown & 9.31E+02 $\pm$(4.35E+01) \tridown \\
C76   & \textbf{1.81E+05 $\pm$(7.92E+03)}\triup & 2.38E+05 $\pm$(1.00E+04) \tridown & 2.05E+05 $\pm$(8.43E+03) \tridown & 2.01E+05 $\pm$(6.79E+03) \tridown & 2.43E+05 $\pm$(4.20E+03) \tridown & 2.08E+05 $\pm$(9.29E+03) \tridown \\
C99   & \textbf{1.65E+03 $\pm$(4.97E+01)}\triup & 2.11E+03 $\pm$(2.40E+01) \tridown & 1.84E+03 $\pm$(3.67E+01) \tridown & 1.84E+03 $\pm$(3.12E+01) \tridown & 2.45E+03 $\pm$(3.83E+01) \tridown & 1.94E+03 $\pm$(5.19E+01) \tridown \\
C198  & \textbf{2.76E+04 $\pm$(2.79E+02)}\triup & 3.61E+04 $\pm$(9.48E+02) \tridown & 3.07E+04 $\pm$(1.38E+03) \tridown & 3.19E+04 $\pm$(5.75E+02) \tridown & 5.06E+04 $\pm$(9.51E+02) \tridown & 3.37E+04 $\pm$(9.22E+02) \tridown \\
C318  & \textbf{5.83E+04 $\pm$(1.17E+03)}\triup & 9.52E+04 $\pm$(4.39E+03) \tridown & 6.48E+04 $\pm$(2.35E+03) \tridown & 7.07E+04 $\pm$(8.70E+02) \tridown & 1.55E+05 $\pm$(4.41E+03) \tridown & 7.34E+04 $\pm$(2.57E+03) \tridown \\
C417  & \textbf{2.42E+04 $\pm$(9.92E+02)}\triup & 3.33E+04 $\pm$(1.64E+03) \tridown & 2.69E+04 $\pm$(9.20E+02) \tridown & 2.77E+04 $\pm$(1.04E+03) \tridown & 9.96E+04 $\pm$(3.25E+03) \tridown & 2.82E+04 $\pm$(3.15E+02) \tridown \\
C442  & \textbf{7.25E+04 $\pm$(1.01E+03)}\triup & 1.24E+05 $\pm$(3.81E+03) \tridown & 8.22E+04 $\pm$(1.27E+03) \tridown & 8.38E+04 $\pm$(2.52E+03) \tridown & 2.02E+05 $\pm$(5.85E+03) \tridown & 7.93E+04 $\pm$(3.35E+03) \tridown \\
C574  & \textbf{6.05E+04 $\pm$(8.40E+02)}\triup & 1.03E+05 $\pm$(3.24E+03) \tridown & 6.49E+04 $\pm$(1.93E+03) \tridown & 6.86E+04 $\pm$(2.12E+03) \tridown & 1.71E+05 $\pm$(1.94E+03) \tridown & 7.27E+04 $\pm$(3.07E+03) \tridown \\
C783  & \textbf{1.21E+04 $\pm$(4.18E+02)}\triup & 2.11E+04 $\pm$(7.97E+02) \tridown & 1.32E+04 $\pm$(4.84E+02) \tridown & 1.42E+04 $\pm$(5.55E+02) \tridown & 4.46E+04 $\pm$(5.40E+02) \tridown & 1.50E+04 $\pm$(1.68E+02) \tridown \\
C1002 & \textbf{3.74E+05 $\pm$(1.59E+04)}\triup & 7.12E+05 $\pm$(3.35E+04) \tridown & 4.29E+05 $\pm$(1.97E+04) \tridown & 4.48E+05 $\pm$(2.13E+04) \tridown & 1.55E+06 $\pm$(4.75E+04) \tridown & 4.70E+05 $\pm$(1.72E+04) \tridown \\
C1060 & \textbf{3.46E+05 $\pm$(9.73E+03)}\triup & 6.49E+05 $\pm$(2.96E+04) \tridown & 3.97E+05 $\pm$(1.07E+04) \tridown & 4.16E+05 $\pm$(1.99E+04) \tridown & 1.53E+06 $\pm$(3.78E+04) \tridown & 4.38E+05 $\pm$(2.07E+04) \tridown \\
C1084 & \textbf{3.62E+05 $\pm$(5.53E+03)}\triup & 6.67E+05 $\pm$(2.35E+04) \tridown & 4.08E+05 $\pm$(1.59E+04) \tridown & 4.32E+05 $\pm$(2.05E+04) \tridown & 1.98E+06 $\pm$(2.91E+04) \tridown & 4.49E+05 $\pm$(1.51E+04) \tridown \\
C1173 & \textbf{8.17E+04 $\pm$(3.32E+03)}\triup & 1.67E+05 $\pm$(6.28E+03) \tridown & 9.36E+04 $\pm$(2.18E+03) \tridown & 1.00E+05 $\pm$(2.57E+03) \tridown & 3.36E+05 $\pm$(1.41E+04) \tridown & 9.70E+04 $\pm$(4.16E+03) \tridown \\
C1291 & \textbf{7.53E+04 $\pm$(3.09E+03)}\triup & 1.51E+05 $\pm$(7.27E+03) \tridown & 8.12E+04 $\pm$(2.74E+03) \tridown & 8.72E+04 $\pm$(2.58E+03) \tridown & 4.13E+05 $\pm$(1.56E+04) \tridown & 8.34E+04 $\pm$(2.66E+03) \tridown \\
C1304 & \textbf{3.86E+05 $\pm$(1.41E+04)}\triup & 7.41E+05 $\pm$(8.15E+03) \tridown & 4.20E+05 $\pm$(2.04E+04) \tridown & 4.47E+05 $\pm$(4.97E+03) \tridown & 2.22E+06 $\pm$(8.20E+04) \tridown & 4.84E+05 $\pm$(2.05E+04) \tridown \\
C1400 & \textbf{3.52E+04 $\pm$(6.72E+02)}\triup & 5.10E+04 $\pm$(1.05E+03) \tridown & 3.88E+04 $\pm$(1.32E+03) \tridown & 3.98E+04 $\pm$(7.97E+02) \tridown & 3.19E+05 $\pm$(7.68E+03) \tridown & 4.18E+04 $\pm$(1.61E+03) \tridown \\
C1432 & \textbf{2.31E+05 $\pm$(4.68E+03)}\triup & 4.69E+05 $\pm$(1.17E+04) \tridown & 2.69E+05 $\pm$(7.67E+03) \tridown & 2.77E+05 $\pm$(5.81E+03) \tridown & 9.93E+05 $\pm$(1.05E+04) \tridown & 2.60E+05 $\pm$(6.94E+03) \tridown \\
C1577 & \textbf{3.82E+04 $\pm$(6.27E+02)}\triup & 6.50E+04 $\pm$(1.70E+03) \tridown & 4.28E+04 $\pm$(1.76E+03) \tridown & 4.38E+04 $\pm$(5.78E+02) \tridown & 2.81E+05 $\pm$(9.86E+03) \tridown & 4.39E+04 $\pm$(6.01E+02) \tridown \\
\bottomrule
\triup/\tridown/\diamondfill & 21/0/0 & 0/21/0 & 0/21/0 & 0/21/0 & 0/21/0 & 0/21/0 \\
\bottomrule
\end{tabular}
}
\end{table*}

\begin{table*}[htbp]
\centering
\caption{Performance Comparison of the Maximum Path Length (mean~$\pm$~std) for the TSPLIB using a team of 8 Robots. Triangle-up (\triup) = significantly better; diamond (\diamondfill) = statistically equal; triangle-down (\tridown) = significantly worse, according to Wilcoxon signed-rank tests versus the best algorithm in each row.  Best mean values are in \textbf{bold}.}
\label{tab:max_rc8}
\scriptsize
\resizebox{\textwidth}{!}{
\begin{tabular}{c|cccccc}
\toprule
\textbf{ID} & \textbf{SINE} & \textbf{ACO} & \textbf{AR-ACO} & \textbf{DL-ACO} & \textbf{IEACO} & \textbf{Smooth ACO} \\
\midrule
C51   & 1.01E+02 $\pm$(2.15E+01) \tridown & 1.01E+02 $\pm$(1.25E+01) \tridown & \textbf{9.28E+01 $\pm$(1.01E+01)}\triup & 9.41E+01 $\pm$(1.19E+01) \tridown & 1.20E+02 $\pm$(1.02E+01) \tridown & 1.07E+02 $\pm$(3.15E+00) \tridown \\
C52   & 2.56E+03 $\pm$(5.58E+02) \tridown & 3.61E+03 $\pm$(7.90E+02) \tridown & \textbf{2.47E+03 $\pm$(2.06E+02)}\triup & 2.68E+03 $\pm$(5.07E+02) \tridown & 3.07E+03 $\pm$(2.39E+01) \tridown & 2.57E+03 $\pm$(3.70E+02) \tridown \\
C70   & \textbf{1.71E+02 $\pm$(8.63E+00)}\triup & 1.84E+02 $\pm$(2.43E+01) \tridown & 1.78E+02 $\pm$(4.15E+01) \tridown & 1.75E+02 $\pm$(1.19E+01) \tridown & 1.88E+02 $\pm$(2.94E+00) \tridown & 1.75E+02 $\pm$(3.85E+01) \tridown \\
C76   & \textbf{2.00E+04 $\pm$(1.73E+04)}\triup & 2.83E+04 $\pm$(2.63E+04) \tridown & 2.26E+04 $\pm$(1.95E+04) \tridown & 2.42E+04 $\pm$(2.10E+04) \tridown & 2.67E+04 $\pm$(2.39E+04) \tridown & 2.20E+04 $\pm$(1.90E+04) \tridown \\
C99   & \textbf{3.44E+02 $\pm$(4.55E+00)}\triup & 3.86E+02 $\pm$(6.15E+00) \tridown & 3.55E+02 $\pm$(4.47E+01) \tridown & 3.59E+02 $\pm$(5.97E+01) \tridown & 5.12E+02 $\pm$(5.60E+00) \tridown & 3.47E+02 $\pm$(5.01E+01) \tridown \\
C198   & 6.58E+03 $\pm$(1.28E+02) \tridown & 6.50E+03 $\pm$(9.47E+01) \tridown & 6.41E+03 $\pm$(8.42E+01) \tridown & \textbf{6.32E+03 $\pm$(5.99E+02)}\triup & 9.45E+03 $\pm$(5.79E+02) \tridown & 6.35E+03 $\pm$(3.30E+02) \tridown \\
C318   & \textbf{1.01E+04 $\pm$(5.93E+02)}\triup & 1.66E+04 $\pm$(6.51E+03) \tridown & 1.28E+04 $\pm$(2.46E+03) \tridown & 1.19E+04 $\pm$(9.42E+02) \tridown & 2.34E+04 $\pm$(2.17E+03) \tridown & 1.49E+04 $\pm$(4.04E+03) \tridown \\
C417   & 5.71E+03 $\pm$(3.13E+02) \tridown & 6.38E+03 $\pm$(3.20E+03) \tridown & 4.69E+03 $\pm$(1.65E+02) \tridown & \textbf{4.41E+03 $\pm$(9.88E+02)}\triup & 1.41E+04 $\pm$(5.77E+03) \tridown & 5.52E+03 $\pm$(1.39E+03) \tridown \\
C442   & \textbf{1.04E+04 $\pm$(4.06E+01)}\triup & 1.78E+04 $\pm$(6.86E+03) \tridown & 1.66E+04 $\pm$(3.62E+03) \tridown & 1.46E+04 $\pm$(1.10E+03) \tridown & 2.92E+04 $\pm$(1.20E+04) \tridown & 1.72E+04 $\pm$(3.50E+03) \tridown \\
C574   & \textbf{8.03E+03 $\pm$(9.70E+02)}\triup & 1.57E+04 $\pm$(3.58E+03) \tridown & 1.26E+04 $\pm$(1.45E+03) \tridown & 1.29E+04 $\pm$(2.93E+03) \tridown & 2.40E+04 $\pm$(1.13E+04) \tridown & 1.39E+04 $\pm$(2.62E+03) \tridown \\
C783   & \textbf{1.75E+03 $\pm$(7.98E-01)}\triup & 3.23E+03 $\pm$(1.10E+03) \tridown & 2.78E+03 $\pm$(7.43E+01) \tridown & 2.48E+03 $\pm$(3.04E+01) \tridown & 6.12E+03 $\pm$(1.64E+03) \tridown & 2.88E+03 $\pm$(1.93E+02) \tridown \\
C1002   & \textbf{5.07E+04 $\pm$(5.49E+03)}\triup & 1.36E+05 $\pm$(4.11E+04) \tridown & 8.20E+04 $\pm$(2.25E+03) \tridown & 9.89E+04 $\pm$(1.23E+04) \tridown & 2.50E+05 $\pm$(1.28E+05) \tridown & 1.01E+05 $\pm$(1.86E+03) \tridown \\
C1060   & \textbf{4.68E+04 $\pm$(3.42E+03)}\triup & 1.19E+05 $\pm$(5.97E+04) \tridown & 8.33E+04 $\pm$(7.22E+03) \tridown & 8.86E+04 $\pm$(1.02E+04) \tridown & 2.35E+05 $\pm$(6.08E+04) \tridown & 9.29E+04 $\pm$(9.51E+03) \tridown \\
C1084   & \textbf{6.42E+04 $\pm$(5.54E+03)}\triup & 1.43E+05 $\pm$(3.58E+04) \tridown & 9.62E+04 $\pm$(5.84E+03) \tridown & 8.48E+04 $\pm$(8.43E+03) \tridown & 2.68E+05 $\pm$(1.38E+05) \tridown & 1.10E+05 $\pm$(4.82E+03) \tridown \\
C1173   & \textbf{1.09E+04 $\pm$(8.51E+02)}\triup & 2.64E+04 $\pm$(1.00E+04) \tridown & 2.07E+04 $\pm$(2.22E+03) \tridown & 2.21E+04 $\pm$(2.81E+01) \tridown & 5.12E+04 $\pm$(2.33E+04) \tridown & 2.43E+04 $\pm$(2.38E+03) \tridown \\
C1291   & \textbf{1.32E+04 $\pm$(1.32E+02)}\triup & 3.96E+04 $\pm$(2.62E+02) \tridown & 1.82E+04 $\pm$(2.19E+02) \tridown & 1.61E+04 $\pm$(1.80E+02) \tridown & 7.55E+04 $\pm$(2.05E+02) \tridown & 1.40E+04 $\pm$(2.08E+02) \tridown \\
C1304   & \textbf{5.52E+04 $\pm$(5.52E+02)}\triup & 1.76E+05 $\pm$(8.11E+02) \tridown & 9.85E+04 $\pm$(7.85E+02) \tridown & 9.33E+04 $\pm$(9.22E+02) \tridown & 4.53E+05 $\pm$(8.05E+02) \tridown & 9.39E+04 $\pm$(9.73E+02) \tridown \\
C1400   & \textbf{7.54E+03 $\pm$(7.54E+01)}\triup & 2.03E+04 $\pm$(1.01E+02) \tridown & 1.02E+04 $\pm$(1.40E+02) \tridown & 1.08E+04 $\pm$(1.04E+02) \tridown & 4.90E+04 $\pm$(1.36E+02) \tridown & 9.46E+03 $\pm$(1.14E+02) \tridown \\
C1432   & \textbf{3.09E+04 $\pm$(3.09E+02)}\triup & 9.38E+04 $\pm$(6.05E+02) \tridown & 5.51E+04 $\pm$(4.72E+02) \tridown & 6.10E+04 $\pm$(5.52E+02) \tridown & 1.71E+05 $\pm$(3.76E+02) \tridown & 5.24E+04 $\pm$(3.79E+02) \tridown \\
C1577   & \textbf{5.78E+03 $\pm$(5.78E+01)}\triup & 1.43E+04 $\pm$(7.46E+01) \tridown & 8.58E+03 $\pm$(1.15E+02) \tridown & 9.33E+03 $\pm$(8.60E+01) \tridown & 4.40E+04 $\pm$(1.04E+02) \tridown & 7.03E+03 $\pm$(7.86E+01) \tridown \\
\bottomrule
\triup/\tridown/\diamondfill & 17/4/0 & 0/21/0 & 2/19/0 & 2/19/0 & 0/21/0 & 0/21/0 \\
\bottomrule
\end{tabular}
}
\end{table*}

To assess the transferability to realistic scenarios, we further evaluate our framework on a coastal sector of Antarctica near the South Pole. Fifty nodes are randomly sampled from the valid land area, and the robots are initialized at fixed locations along the coast. We isolate routing behavior with a total distance cost. The compared methods follow those in our main study: SINE, ACO, AR-ACO, DL-ACO, and IE-ACO. The results are averaged over 10 independent runs for each configuration. We vary the robots with 2, 4, 6, and 8 to form a grid of scenarios. Table~\ref{tab:southpole-results} summarizes the results; showing that SINE achieves superior and scalable result regardless of the robot configuration. On average, SINE achieves the lowest total path length while recording the highest number of per configuration wins. A graphical illustration of the node distribution and path connection for each robot is presented in Fig.~\ref{fig:southpole-combined}.

\begin{table*}[htbp]
\centering
\caption{Performance Comparison of the Total Path Length (mean~$\pm$~std) for the Antarctica scenario of 50 nodes using different robotic teams. Triangle-up (\triup) = significantly better; diamond (\diamondfill) = statistically equal; triangle-down (\tridown) = significantly worse, according to Wilcoxon signed-rank tests versus the best algorithm in each row.  Best mean values are in \textbf{bold}.}
\label{tab:southpole-results}
\scriptsize
\resizebox{\textwidth}{!}{
\begin{tabular}{c|ccccc}
\toprule
\# \textbf{Robots} & \textbf{SINE} & \textbf{ACO} & \textbf{AR-ACO} & \textbf{DL-ACO} & \textbf{IE-ACO} \\
\hline
2 & 1.47E+04  $\pm$(6.25E+03) \triup & 1.51E+04  $\pm$(6.60E+03) \tridown & 1.85E+04  $\pm$(8.53E+03) \tridown & 1.50E+04  $\pm$(6.45E+03) \tridown & 1.53E+04  $\pm$(6.77E+03) \tridown \\
4 & 1.69E+04  $\pm$(5.83E+03) \triup & 1.72E+04   $\pm$(6.20E+03) \tridown & 1.94E+04  $\pm$(6.76E+03) \tridown & 1.72E+04  $\pm$(6.13E+03) \tridown & 1.79E+04  $\pm$(6.87E+03) \tridown \\
6 & 1.96E+04  $\pm$(6.61E+03) \triup & 1.97E+04  $\pm$(6.80E+03) \tridown & 2.20E+04  $\pm$(1.07E+04) \tridown & 1.99E+04  $\pm$(7.11E+03) \tridown & 1.98E+04  $\pm$(7.21E+03) \tridown \\
8 & 2.18E+04  $\pm$(6.74E+03) \triup & 2.21E+04  $\pm$(7.15E+03) \tridown & 2.46E+04  $\pm$(1.37E+04) \tridown & 2.20E+04  $\pm$(7.13E+03) \tridown & 2.21E+04  $\pm$(7.26E+03) \tridown \\
\bottomrule
\triup/\tridown/\diamondfill & 4/0/0 & 0/4/0 & 0/4/0 & 0/4/0 & 0/4/0  \\
\bottomrule
\end{tabular}
}
\end{table*}

\begin{table*}[htbp]
\centering
\caption{Performance Comparison of the Maximum Path Length (mean~$\pm$~std) for the Antarctica scenario of 50 nodes using different robotic teams. Triangle-up (\triup) = significantly better; diamond (\diamondfill) = statistically equal; triangle-down (\tridown) = significantly worse, according to Wilcoxon signed-rank tests versus the best algorithm in each row. Best mean values are in \textbf{bold}.}
\label{tab:max_southpole-results}
\scriptsize
\resizebox{\textwidth}{!}{
\begin{tabular}{c|ccccc}
\toprule
\# \textbf{Robots} & \textbf{SINE} & \textbf{ACO} & \textbf{AR-ACO} & \textbf{DL-ACO} & \textbf{IEACO} \\
\hline
2 & \textbf{1.13E+04}~$\pm$(4.05E+02)~\triup & 1.19E+04~$\pm$(7.10E+02)~\tridown & 2.02E+04~$\pm$(1.29E+04)~\tridown & 1.15E+04~$\pm$(3.11E+02)~\tridown & 1.39E+04~$\pm$(4.09E+02)~\tridown \\
4 & \textbf{7.03E+03~$\pm$(5.88E+02)}~\triup & 7.55E+03~$\pm$(1.19E+03)~\tridown & 7.75E+03~$\pm$(6.36E+02)~\tridown & 7.212E+03~$\pm$(8.08E+02)~\tridown & 7.63E+03~$\pm$(8.83E+02)~\tridown \\
6 & \textbf{6.10E+03}~$\pm$(2.15E+02)~\triup & 6.29E+03~$\pm$(2.41E+02)~\tridown & 6.28E+03~$\pm$(4.20E+02)~\tridown & 6.19E+03~$\pm$(2.80E+02)~\tridown & 6.40E+03~$\pm$(5.50E+02)~\tridown \\
8 & 5.42E+03~$\pm$(1.80E+02)~\tridown & 5.55E+03~$\pm$(2.25E+02)~\tridown & \textbf{5.15E+03}~$\pm$(5.27E+02)~\triup & 5.49E+03~$\pm$(2.09E+02)~\tridown & 5.41E+03~$\pm$(7.11E+02)~\tridown \\
\bottomrule
\triup/\tridown/\diamondfill & 3/1/0 & 0/4/0 & 3/1/0 & 0/4/1 & 0/4/0 \\
\bottomrule
\end{tabular}
}
\end{table*}

\subsection{Multiple Problem Performance Analysis}
To further explore the significance of the proposed algorithm's performance, we present a comprehensive multi-problem analysis using the non-parametric Friedman ranking with Holm correction. This analysis was conducted in stages of a team of 2, 4, 6, 8 robots, and a combined analysis involving all the results. Table \ref{tab:friedman_ranking_combined} shows the result of the Friedman Ranking analysis using the total path length. In addition, a complementary assessment in Table \ref{tab:max_friedman_ranking_combined} focuses on the maximum single-robot tour length. 


The results of the Friedman ranking test demonstrate that SINE outperforms baseline methods in all cases, showing a clear significant statistical performance in both total path length and maximum length as shown in Tables \ref{tab:friedman_ranking_combined} and \ref{tab:max_friedman_ranking_combined}, respectively. Its scalability is clear, with benefits becoming more noticeable as the problem size grows, while maintaining computational efficiency thanks to the targeted search approach. The results of these analyses are in consonance with the single problem analysis; SINE consistently achieves the lowest average ranks, indicating better route compactness, less variance, and improved coordination among robots compared to baselines such as ACO, AR-ACO, DL-ACO, IEACO, and Smooth ACO. This superiority stems from the integration of a minimum spanning tree backbone that guides exploration and ensures scalability in large scale instances. The AR-ACO ranks second in comparison to our proposed algorithm, leveraging the assisted A* repulsive field rules to guide the search process. And closely following AR-ACO is the DL-ACO, which presents a double layer optimization mechanism. This indicates that effectively steering the ACO search process consistently results in superior performance.

\begin{table}[htbp]
    \centering
    \caption{Friedman Ranking of each algorithm performance across Robot Counts (lower is better)}
    \label{tab:friedman_ranking_combined}
    \resizebox{\textwidth}{!}{
    \begin{tabular}{lcccccccc}
    \hline
    \multirow{2}{*}{\textbf{Algorithms}} & \multicolumn{2}{c}{\textbf{2 Robots}} & \multicolumn{2}{c}{\textbf{4 Robots}} &  \multicolumn{2}{c}{\textbf{8 Robots}} & \multicolumn{2}{c}{\textbf{Overall}} \\
    \cmidrule(lr){2-9}
    & Friedman Score & Rank & Friedman Score & Rank & Friedman Score & Rank & Friedman Score & Rank \\
    \hline
    SINE        & 1.12 & 1 & 1.15 & 1 & 1.23 & 1  & 1.17 & 1  \\
    AR-ACO      & 3.08 & 3 & 2.58  & 2 &  2.62 & 2  & 2.73 & 2  \\
    DL-ACO      & 2.64 & 2 & 2.73 & 3 &  3.04 & 3 & 2.80 & 3 \\
    Smooth ACO  & 4.48 & 5 & 4.12 & 4 &  4.00 & 4 & 4.17 & 4 \\
    ACO         & 3.84 & 4 & 4.50 & 5 &  4.58 & 5 & 4.34 & 5 \\
    IEACO       & 5.84 & 6 & 5.92 & 6 &  5.54 & 6 & 5.78 & 6 \\
    \hline
    \end{tabular}
    }
\end{table}

\begin{table}[htbp]
    \centering
    \caption{Friedman Ranking of Maximum Length performance across Robot Counts (lower is better)}
    \label{tab:max_friedman_ranking_combined}
    \resizebox{\textwidth}{!}{
    \begin{tabular}{lcccccccc}
    \hline
    \multirow{2}{*}{\textbf{Algorithms}} & \multicolumn{2}{c}{\textbf{2 Robots}} & \multicolumn{2}{c}{\textbf{4 Robots}} &  \multicolumn{2}{c}{\textbf{8 Robots}} & \multicolumn{2}{c}{\textbf{Overall}} \\
    \cmidrule(lr){2-9}
    & Friedman Score & Rank & Friedman Score & Rank & Friedman Score & Rank & Friedman Score & Rank \\
    \hline
    SINE        & 1.35 & 1 & 1.21 & 1 & 1.53 & 1  & 1.36 & 1  \\
    AR-ACO      & 2.40 & 2 & 2.60 & 2 & 2.70 & 2  & 2.57 & 2  \\
    DL-ACO      & 2.83 & 3 & 2.60 & 3 & 2.78 & 3  & 2.73 & 3 \\
    Smooth ACO  & 3.58 & 4 & 3.74 & 4 & 3.13 & 4  & 3.48 & 4  \\
    ACO         & 4.85 & 5 & 4.90 & 5 & 4.98 & 5  & 4.91 & 5  \\
    IEACO       & 6.00 & 6 & 5.95 & 6 & 5.90 & 6  & 5.95 & 6 \\
    \hline
    \end{tabular}
    }
\end{table}

\subsection{Ablation Study}
We conducted ablation experiments to evaluate how varying the structural prior weight \(\alpha\) impacts the total path length across different problem scales and fleet sizes. Table~\ref{tab:ablation} presents the total path lengths for each combination, with lower values indicating improved route quality. The results reveal that the selection of \(\alpha\) significantly affects routing performance: in smaller scenarios with fewer robots, higher \(\alpha\) values achieve the shortest paths. As the problem scale increases, higher \(\alpha\) settings consistently produce the most efficient path lengths across various configurations. And non-monotonic trends are noticeable, where lower \(\alpha\) values outperform higher ones in some cases. Overall, these findings suggest that a carefully selected prior bias can greatly enhance solution quality, while either too little or too much bias may reduce routing efficiency.

As the number of robots or city count increases, SINE tends to outperform the baseline across most \(\alpha\) settings, with the best performing weights often corresponding to the lowest values in each category. Although the structural prior has a limited impact in very small scale scenarios, its influence becomes more pronounced as task complexity grows. Introducing the backbone helps stabilize path length reductions and improves load balancing, confirming the effectiveness of this approach in handling more intricate multi robot path planning challenges.

\begin{table}[htbp]
\centering
\caption{Effect of structural prior weight $\alpha$ on total and max single robot path lengths.}
\label{tab:ablation}
\resizebox{0.5\textwidth}{!}{%
\begin{tabular}{c ccc}
\toprule
\multicolumn{4}{c}{\textbf{50 cities}} \\
\cmidrule(lr){1-4}
$\alpha$ & 2 robots & 4 robots & 8 robots \\
\midrule
\multicolumn{4}{l}{\textbf{Total path length}} \\
\midrule
0.0 & 5.76e+02   $\pm$ (2.10e+01) & 6.85e+02  $\pm$ (8.00e+00) & 7.73e+02  $\pm$ (2.80e+01) \\
0.1 & 5.75e+02  $\pm$ (6.00e+00) & \textbf{6.30e+02  $\pm$ (2.60e+01)} & \textbf{7.31e+02  $\pm$ (2.40e+01)} \\
0.5 & 6.00e+02  $\pm$ (1.30e+01) & 6.71e+02  $\pm$ (2.80e+01) & 7.45e+02  $\pm$ (1.60e+01) \\
1.0 & 5.83e+02  $\pm$ (1.10e+01) & 6.79e+02  $\pm$ (2.30e+01) & 7.56e+02  $\pm$ (2.50e+01) \\
2.0 & 5.80e+02  $\pm$ (2.30e+01) & 6.53e+02  $\pm$ (1.00e+01) & 7.60e+02  $\pm$ (3.30e+01) \\
5.0 & 6.05e+02  $\pm$ (2.20e+01) & 6.77e+02  $\pm$ (2.90e+01) & 7.52e+02  $\pm$ (1.20e+01) \\
10.0 & \textbf{5.75e+02  $\pm$ (2.60e+01)} & 6.48e+02  $\pm$ (1.00e+01) & 7.53e+02  $\pm$ (3.30e+01) \\
\midrule
\multicolumn{4}{l}{\textbf{Max single robot path length}} \\
\midrule
0.0 & \textbf{3.06e+02  $\pm$ (1.10e+01)} & 2.12e+02  $\pm$ (2.00e+00) & 1.59e+02  $\pm$ (6.00e+00) \\
0.1 & \textbf{3.06e+02  $\pm$ (3.00e+00)} & 1.92e+02  $\pm$ (8.00e+00) & 1.27e+02  $\pm$ (4.00e+00) \\
0.5 & 3.11e+02  $\pm$ (7.00e+00) & 1.97e+02  $\pm$ (8.00e+00) & 1.27e+02  $\pm$ (3.00e+00) \\
1.0 & 3.19e+02  $\pm$ (6.00e+00) & \textbf{1.77e+02  $\pm$ (6.00e+00)} & \textbf{1.23e+02  $\pm$ (4.00e+00)} \\
2.0 & \textbf{3.06e+02  $\pm$ (1.20e+01)} & 1.92e+02  $\pm$ (3.00e+00) & 1.27e+02  $\pm$ (6.00e+00) \\
5.0 & 3.11e+02  $\pm$ (1.20e+01) & 1.97e+02  $\pm$ (3.00e+00) & 1.27e+02  $\pm$ (6.00e+00) \\
10.0 & \textbf{3.06e+02  $\pm$ (1.40e+01)} & 1.92e+02  $\pm$ (9.00e+00) & 1.27e+02  $\pm$ (6.00e+00) \\
\bottomrule
\end{tabular}
}

\resizebox{0.5\textwidth}{!}{%
\begin{tabular}{c ccc}
\multicolumn{4}{c}{\textbf{100 cities}} \\
\cmidrule(lr){1-4}
$\alpha$ & 2 robots & 4 robots & 8 robots \\
\midrule
\multicolumn{4}{l}{\textbf{Total path length}} \\
\midrule
0.0 & \textbf{7.72e+02  $\pm$ (3.50e+01)} & 9.05e+02  $\pm$ (1.00e+01) & 9.99e+02  $\pm$ (2.50e+01) \\
0.1 & 8.03e+02  $\pm$ (3.80e+01) & 9.03e+02  $\pm$ (3.20e+01) & 1.004e+03  $\pm$ (3.30e+01) \\
0.5 & 8.22e+02  $\pm$ (3.10e+01) & 9.30e+02  $\pm$ (1.90e+01) & 1.046e+03  $\pm$ (4.60e+01) \\
1.0 & 7.97e+02  $\pm$ (9.00e+00) & 9.05e+02  $\pm$ (3.20e+01) & 9.99e+02  $\pm$ (3.30e+01) \\
2.0 & 7.95e+02  $\pm$ (2.90e+01) & 9.06e+02  $\pm$ (1.90e+01) & 9.98e+02  $\pm$ (4.20e+01) \\
5.0 & 8.20e+02  $\pm$ (1.00e+01) & 9.21e+02  $\pm$ (3.20e+01) & 1.026e+03  $\pm$ (3.40e+01) \\
10.0 & 7.90e+02  $\pm$ (3.60e+01) & \textbf{8.89e+02  $\pm$ (4.10e+01)} & \textbf{9.91e+02  $\pm$ (1.50e+01)} \\
\midrule
\multicolumn{4}{l}{\textbf{Max single robot path length}} \\
\midrule
0.0 & 4.08e+02  $\pm$ (1.90e+01) & \textbf{2.31e+02  $\pm$ (1.10e+01)} & 1.72e+02  $\pm$ (4.00e+00) \\
0.1 & \textbf{4.04e+02  $\pm$ (1.90e+01)} & 2.69e+02  $\pm$ (9.00e+00) & 1.52e+02  $\pm$ (5.00e+00) \\
0.5 & 4.31e+02  $\pm$ (1.60e+01) & 2.80e+02  $\pm$ (6.00e+00) & 1.74e+02  $\pm$ (8.00e+00) \\
1.0 & 4.16e+02  $\pm$ (5.00e+00) & 2.50e+02  $\pm$ (9.00e+00) & \textbf{1.51e+02  $\pm$ (5.00e+00)} \\
2.0 & 4.13e+02  $\pm$ (1.50e+01) & 2.55e+02  $\pm$ (5.00e+00) & 1.54e+02  $\pm$ (6.00e+00) \\
5.0 & 4.44e+02  $\pm$ (5.00e+00) & 2.81e+02  $\pm$ (1.00e+01) & 1.89e+02  $\pm$ (6.00e+00) \\
10.0 & 4.08e+02  $\pm$ (1.90e+01) & 2.55e+02  $\pm$ (1.20e+01) & 1.54e+02  $\pm$ (2.00e+00) \\
\bottomrule
\end{tabular}
}

\resizebox{0.5\textwidth}{!}{%
\begin{tabular}{c ccc}
\multicolumn{4}{c}{\textbf{200 cities}} \\
\cmidrule(lr){1-4}
$\alpha$ & 2 robots & 4 robots & 8 robots \\
\midrule
\multicolumn{4}{l}{\textbf{Total path length}} \\
\midrule
0.0 & 1.201e+03  $\pm$ (4.90e+01) & 1.263e+03  $\pm$ (3.40e+01) & 1.408e+03  $\pm$ (4.60e+01) \\
0.1 & 1.169e+03  $\pm$ (4.20e+01) & \textbf{1.187e+03  $\pm$ (1.80e+01)} & 1.361e+03  $\pm$ (5.90e+01) \\
0.5 & \textbf{1.144e+03  $\pm$ (4.10e+01)} & 1.204e+03  $\pm$ (5.20e+01) & \textbf{1.355e+03  $\pm$ (2.10e+01)} \\
1.0 & 1.209e+03  $\pm$ (4.40e+01) & 1.316e+03  $\pm$ (2.00e+01) & 1.408e+03  $\pm$ (6.10e+01) \\
2.0 & 1.218e+03  $\pm$ (2.60e+01) & 1.236e+03  $\pm$ (5.20e+01) & 1.408e+03  $\pm$ (2.20e+01) \\
5.0 & 1.214e+03  $\pm$ (4.60e+01) & 1.278e+03  $\pm$ (3.90e+01) & 1.413e+03  $\pm$ (6.10e+01) \\
10.0 & 1.166e+03  $\pm$ (4.20e+01) & 1.219e+03  $\pm$ (1.90e+01) & 1.399e+03  $\pm$ (6.10e+01) \\
\midrule
\multicolumn{4}{l}{\textbf{Max single robot path length}} \\
\midrule
0.0 & 6.19e+02  $\pm$ (2.50e+01) & 3.73e+02  $\pm$ (1.00e+01) & 2.23e+02  $\pm$ (7.00e+00) \\
0.1 & 6.32e+02  $\pm$ (2.30e+01) & \textbf{3.23e+02  $\pm$ (5.00e+00)} & 2.07e+02  $\pm$ (9.00e+00) \\
0.5 & \textbf{5.81e+02  $\pm$ (2.10e+01)} & 3.43e+02  $\pm$ (1.50e+01) & \textbf{2.05e+02  $\pm$ (3.00e+00)} \\
1.0 & 6.76e+02  $\pm$ (2.40e+01) & 3.80e+02  $\pm$ (6.00e+00) & 2.54e+02  $\pm$ (1.10e+01) \\
2.0 & 6.86e+02  $\pm$ (1.40e+01) & 3.72e+02  $\pm$ (1.60e+01) & 2.24e+02  $\pm$ (3.00e+00) \\
5.0 & 6.94e+02  $\pm$ (2.60e+01) & 4.00e+02  $\pm$ (1.20e+01) & 2.24e+02  $\pm$ (1.00e+01) \\
10.0 & 6.84e+02  $\pm$ (2.50e+01) & 3.92e+02  $\pm$ (6.00e+00) & 2.91e+02  $\pm$ (1.30e+01) \\
\bottomrule
\end{tabular}
}
\end{table}


\section{Conclusion}
The comprehensive experimental analysis underscores the effectiveness of SINE in addressing multi robot path planning challenges. The method excels in solution quality, achieving routes with significantly shorter total and maximum single robot path lengths, while maintaining stability across diverse problem scales. Its convergence speed is notably enhanced through the structural prior backbone, which reduces the search space and optimizes computational efficiency compared to traditional ACO variants. Furthermore, SINE demonstrates robustness and reliability, with low result variability and consistent performance under varying conditions, as validated by statistical tests. The innovative integration of a persistent graph skeleton with pheromone based search, coupled with effective multi robot coordination, positions SINE as a strong solution for real time, high quality path planning tasks. We recommend its adoption as the preferred algorithm, particularly for scenarios demanding rapid convergence and scalability, with tailored parameter adjustments to maximize performance.

\section{Future Work}
To further advance SINE, future research should focus on optimizing backbone construction strategies to enhance performance in ultra large scale problems. Developing adaptive parameter adjustment mechanisms could improve the method’s flexibility across diverse scenarios. Additionally, exploring hybrid approaches by integrating SINE with other optimization techniques may yield further performance gains. These directions aim to broaden the algorithm’s applicability and robustness in complex, real world multi-robot path planning environments.

\section*{Acknowlagement}
This work was supported in part by the Korea Institute of Marine Science \& Technology Promotion (KIMST) funded by the Ministry of Trade, Industry and Energy in 2023 (Project Number 20210630), and in part by the MSIT(Ministry of Science and ICT), Korea, under the ITRC (Information Technology Research Center) support program(IITP-2025-RS-2024-00437190, 50 \%) supervised by the IITP(Institute for Information \& Communications Technology Planning \& Evaluation).

\bibliographystyle{cas-model2-names}
\bibliography{cas-refs}


\section*{Appendix}
\appendix
\renewcommand{\thefigure}{A\arabic{figure}}
\setcounter{figure}{0}

\label{appendix:experiment-details}

This appendix provides supplementary visualizations and detailed descriptions of the experiments conducted in Section~\ref{sec:Performance Analysis}. These figures offer deeper insights into the spatial characteristics of robot trajectories produced by various algorithms, as well as the behavior of SINE in diverse urban scenarios.

\begin{figure}[htbp]
 \centering
 \includegraphics[width=0.75\linewidth]{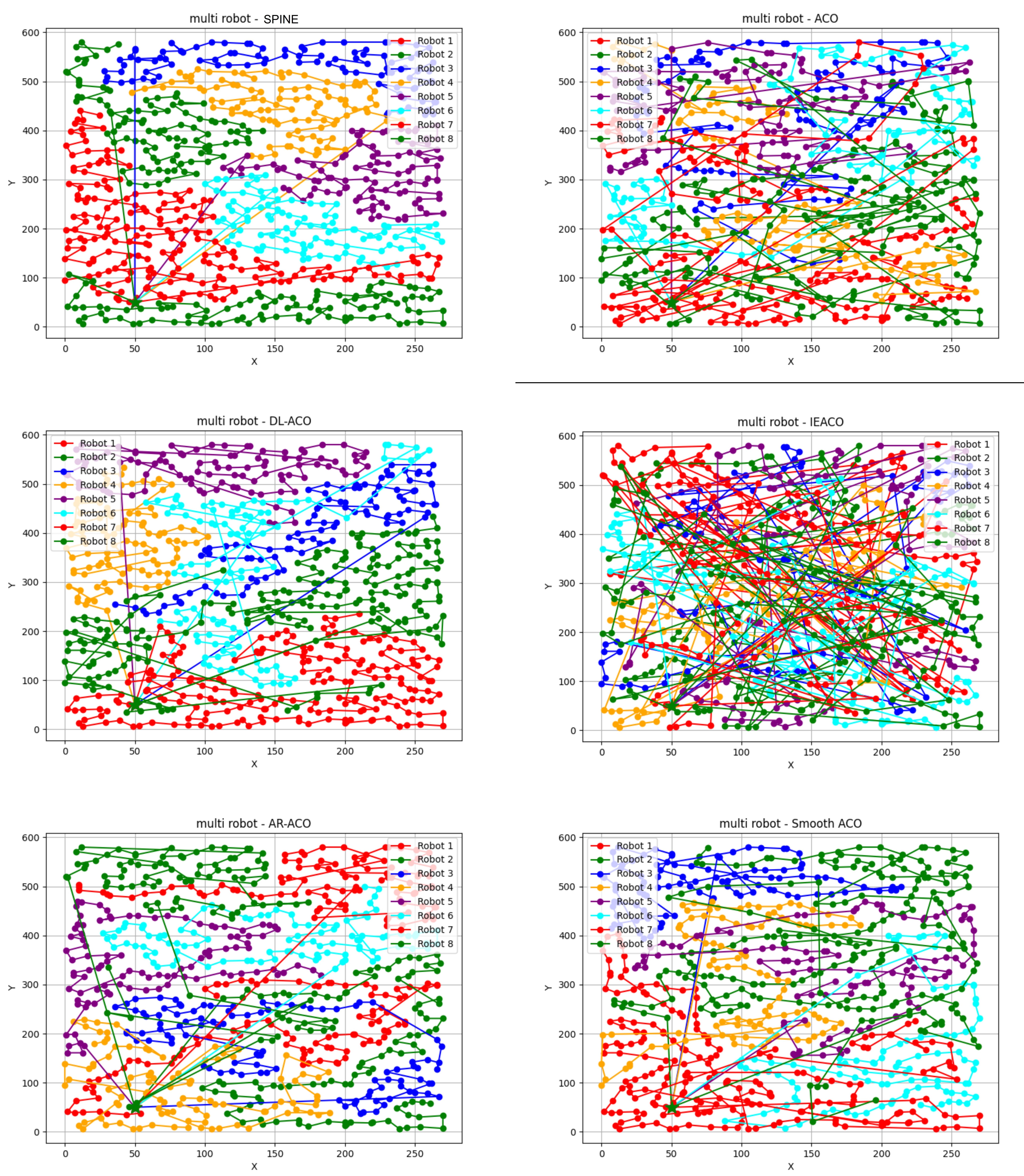}
 \caption{Spatial distribution of robot trajectories for eight different multi-robot routing algorithms on a 200-city, 8-robot instance. From left to right, top to bottom: SINE, ACO, DL-ACO, Genetic Algorithms, AR-ACO, Smooth ACO, IEACO, and Simulated Annealing. SINE demonstrates highly structured and nonoverlapping routes that adhere to the underlying structural backbone. In contrast, other algorithms tend to produce denser, overlapping, or tangled trajectories, which may lead to suboptimal load balancing and increased total travel cost. These differences highlight the effectiveness of structural-prior initialization in producing well organized routes and ensuring clear division of workload among robots.}
 \label{fig:algorithm-distribution}
\end{figure}

\begin{figure}[htbp]
 \centering
 \includegraphics[width=0.92\linewidth]{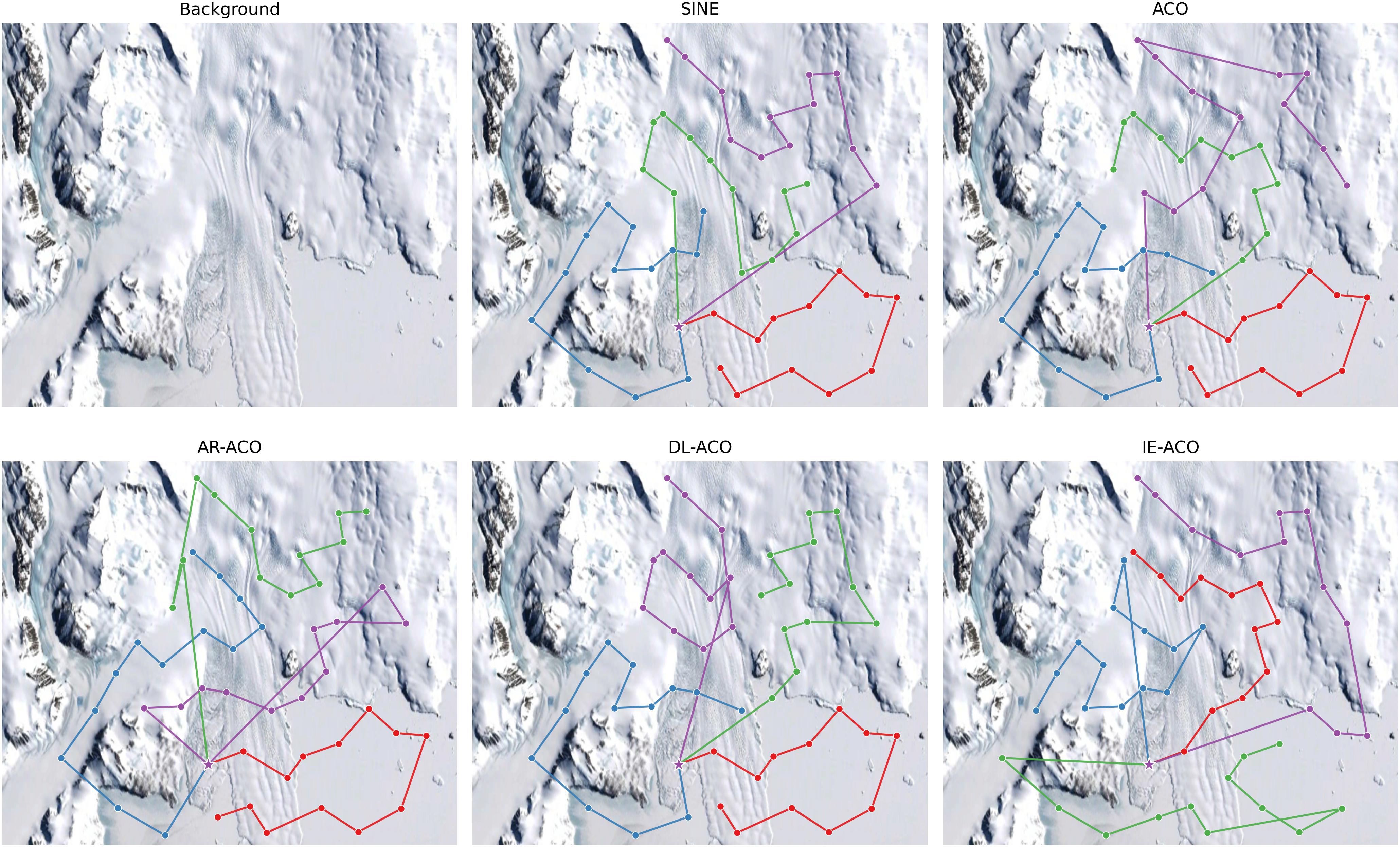}
 \caption{Validation on a South Pole coastal sector in 50 nodes. 
  Pink dots denote target sites sampled from the satellite image, while colored polylines represent robot trajectories produced by different algorithms with SINE, ACO, AR-ACO, DL-ACO, IE-ACO. 
  This experiment evaluates transferability under realistic spatial constraints such as irregular terrain boundaries, clustered node distributions, and heterogeneous densities. 
  Compared to other baselines, SINE produces more compact and balanced routes with reduced overlap, demonstrating its robustness in real world geographic scenarios.}
\label{fig:southpole-combined}
\end{figure}

Across the South Pole sweep we evaluate $5$ node counts $\times$ $4$ robot team sizes, with $6$ independent repeats per combination. Table~\ref{tab:southpole-results} shows that SINE achieves the lowest mean total pixel length for every robot setting (14.7k, 16.9k, 19.6k, and 21.8k pixels for 2, 4, 6, and 8 robots, respectively) together with consistently tight min--max ranges. Figure~\ref{fig:southpole-combined} visualizes a representative overlay: SINE’s tours remain well separated and evenly partition the region, whereas alternative methods display overlapping, tortuous paths.

\begin{figure}[htbp]
 \centering
 \includegraphics[width=0.75\linewidth]{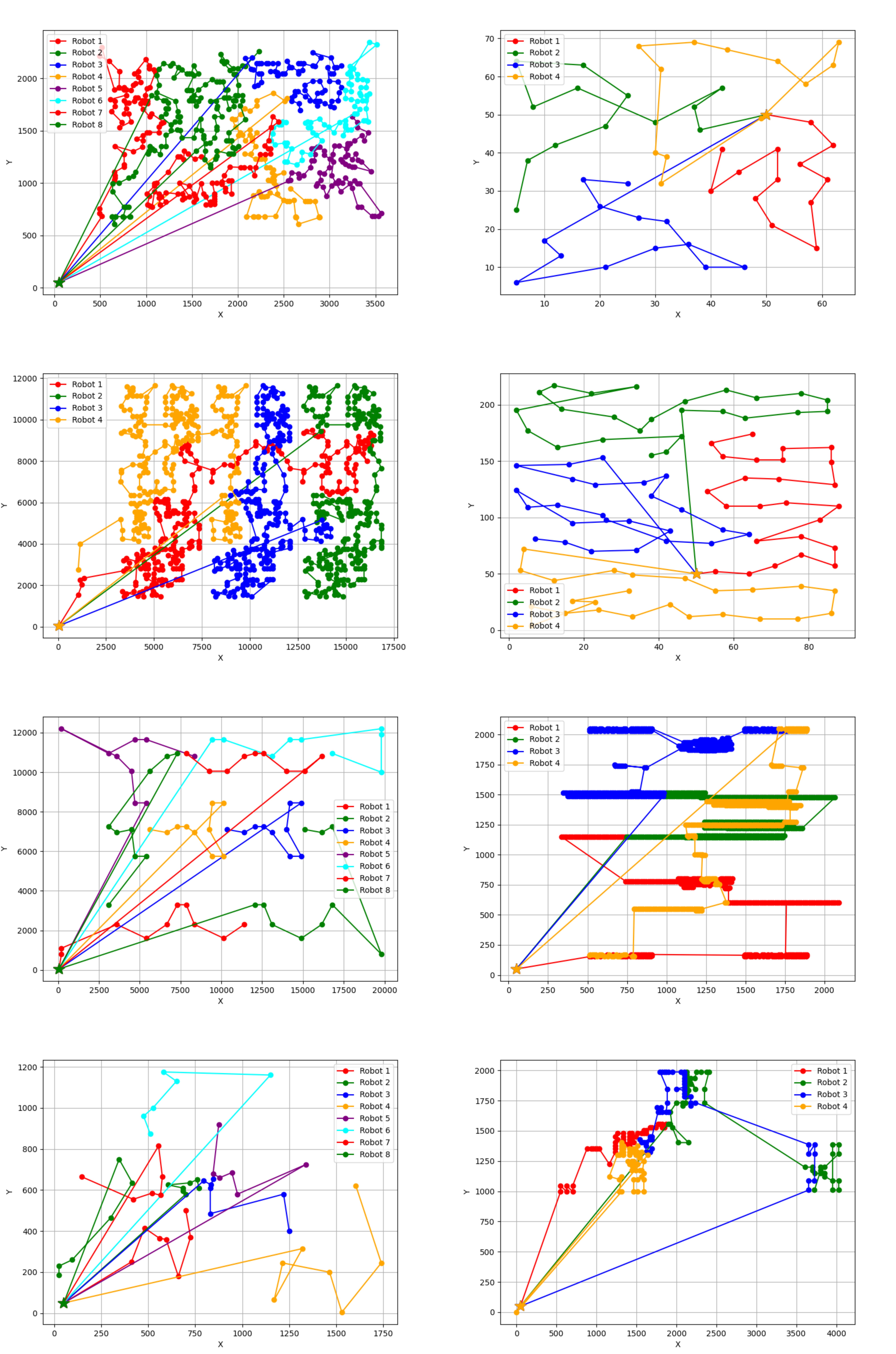}
 \caption{Four representative views of the SINE solution in distinct urban scenarios involving varying numbers of robots (4 or 8) and different city distributions. Each subfigure demonstrates how SINE consistently generates coherent, balanced routing patterns under different scales and spatial layouts. The trajectories reveal how the structural-prior initialization guides robot movement across diverse environments, effectively adapting to variations in city density, regional clustering, and geographical complexity. The algorithm’s ability to preserve load balance and minimize path overlap across different scenarios further verifies its robustness and scalability.}
 \label{fig:mstacoindifferentmaps}
\end{figure}

In Figure~\ref{fig:algorithm-distribution}, we present a comparative analysis of robot trajectory distributions produced by eight distinct algorithms under identical problem settings. Among them, SINE clearly stands out by generating well organized, nonoverlapping routes that closely follow the underlying graph structure, resulting in minimal trajectory crossings and clear division of robot tasks. In contrast, heuristic-based methods such as ACO, DL-ACO, and IE-ACO exhibit more disordered and entangled paths due to their stochastic search mechanisms and absence of structural initialization, often leading to overlapping routes and less efficient coverage. Figure~\ref{fig:mstacoindifferentmaps} further highlights the adaptability of SINE across diverse urban scenarios with varying robot team sizes and city distributions. The visualizations in Figure A2 demonstrate that SINE consistently preserves coherent and balanced routing patterns, even in highly heterogeneous environments with complex clustering, irregular spatial layouts, or significant variations in city density. These results confirm the robustness and scalability of the structural-prior initialization strategy, which effectively guides the robot teams to produce efficient, balanced solutions across a wide range of deployment conditions.













\end{document}